\pdfoutput=1
\documentclass[11pt,letterpaper]{article}

\usepackage[letterpaper,margin=1in,headheight=14pt,headsep=12pt]{geometry}
\usepackage{times}
\usepackage{natbib}
\setcitestyle{authoryear,round,citesep={;},aysep={,},yysep={;}}
\usepackage{fancyhdr}
\usepackage{xcolor}
\usepackage[font=small,labelfont=bf,labelsep=period]{caption}
\usepackage{microtype}
\usepackage[skip=6pt plus 1pt]{parskip}
\usepackage{graphicx}
\usepackage{booktabs}
\usepackage{float}
\usepackage{amsmath}
\usepackage{amssymb}
\usepackage{xspace}
\usepackage{multirow}
\usepackage{tikz}
\usetikzlibrary{arrows.meta,positioning,fit,calc}
\usepackage{hyperref}

\usepackage{algorithm}
\usepackage{algorithmic}

\definecolor{mydarkblue}{rgb}{0,0.08,0.45}
\hypersetup{pdftitle={Hybrid Spark: Adaptive Speculative Decoding That Lets the
  Content Choose the Drafter}, pdfauthor={Ekagra Ranjan, Bharat Venkitesh},
  pdfsubject={}, pdfkeywords={}, pdfborder=0 0 0, colorlinks=true,
  linkcolor=mydarkblue, citecolor=mydarkblue, filecolor=mydarkblue,
  urlcolor=mydarkblue}

\usepackage[capitalize,noabbrev]{cleveref}

\newcommand{\sys}{Hybrid Spark\xspace}
\newcommand{\dspark}{DSpark\xspace}
\newcommand{\av}{AV\xspace}
\newcommand{\x}{$\times$\xspace}

\newcommand{\obs}[1]{\paragraph{Observation #1.}}
\makeatletter
\newcommand{\lnlabel}[1]{\protected@edef\@currentlabel{\arabic{ALC@line}}\label{#1}}
\makeatother

\crefname{algorithm}{Algorithm}{Algorithms}
\Crefname{algorithm}{Algorithm}{Algorithms}
\crefname{ALC@line}{line}{lines}
\Crefname{ALC@line}{Line}{Lines}

\renewenvironment{abstract}
  {\centerline{\large\bfseries Abstract}\vspace{-0.08in}\begin{quote}}
  {\par\end{quote}\vskip 0.12in}

\begin{document}

\begin{center}
{\LARGE\bfseries Hybrid Spark: Adaptive Speculative Decoding\\[0.15em]
That Lets the Content Choose the Drafter\par}
\vspace{1.2em}
{\large Ekagra Ranjan \qquad Bharat Venkitesh\par}
\vspace{0.4em}
Cohere\par
\vspace{0.2em}
\texttt{ekagra@cohere.com}, \texttt{bharat@cohere.com}\par
\end{center}

\vskip 0.2in

\begin{abstract}
Speculative decoding (SD) speeds up large language model (LLM) serving, but its
gain depends on the workload's \emph{shape}, such as batch size and context
length, and on its \emph{content}. A model-based drafter proposes a short block
on any content, while a model-free drafter such as n-gram lookup proposes long
copies at almost no cost when the output repeats its input, as in code editing.
Adaptive verification spends a batch-wide verification budget on the draft
tokens most likely to be accepted, but it supports only one model-based
drafter. Model-free drafters provide no per-token confidence, so they cannot
take part in budget allocation, and deployments must either fix one drafter or
split their GPUs between drafters. We present \sys, a framework that runs a
model-based and a model-free drafter on every step and gives each request the
better of their chains, within adaptive verification and entirely on the GPU,
which keeps asynchronous scheduling intact. The key idea is a price per
verified draft token: the goodput that slot would earn elsewhere in the batch.
Once the model-free drafter's tokens carry calibrated acceptance estimates, fit
once on general text, this price puts both drafters' chains on one scale.
Instantiated with \dspark and an n-gram drafter in vLLM, \sys achieves
1.10--1.62\x the output throughput of \dspark on code editing at batch size~4
and 1.04--1.27\x at batch size~128, while
staying at 0.97--1.01\x of it on chat at temperature~0. A single deployment can
therefore serve both workloads.
\end{abstract}
\thispagestyle{plain}

\section{Introduction}
\label{sec:intro}

Decoding dominates the cost of serving large language models (LLMs). At each step, the model reads its weights and KV cache to produce one token
per request, leaving much of the available compute idle in memory-bound
regimes~\citep{pope2023efficiently}. Speculative decoding
(SD)~\citep{leviathan2023fast,chen2023accelerating} exploits this idle
compute: a cheap drafter proposes several tokens, and the target model
verifies them in a single forward pass while preserving the output
distribution. The benefit depends on how many tokens are accepted relative
to the cost of drafting and verification.

This tradeoff changes with both batch size and content. At small batch
sizes, verifying additional tokens is relatively cheap, and longer drafts
can substantially improve throughput. As the batch grows, verification
becomes more expensive, and verifying too many tokens can make SD slower
than vanilla decoding~\citep{liu2026illusion,liu2024turbospec}.
Adaptive verification, as in TurboSpec~\citep{liu2024turbospec},
\dspark~\citep{cheng2026dspark}, and D-cut~\citep{liu2026dcut},
addresses this problem by adjusting the verification budget at each
step and allocating it to draft tokens most likely to be accepted.
The budget also responds to content: low-entropy domains
such as coding and math tend to accept longer drafts than roleplay
or creative writing~\citep{abramovich2026speedbench}.

Yet adaptive verification assumes that the drafter has already been
chosen. It decides how much of a draft to verify, but not which
drafter should produce it. This distinction matters because different
drafters exploit different kinds of predictability. Model-based
drafters learn to predict the target's continuation and work across
general text. Model-free drafters instead exploit repetition, copying
tokens from the prompt or previously generated text. The best drafter therefore depends
not only on how predictable the next tokens are, but also on where
that predictability comes from.

\begin{figure}[t]
  \centering
  \includegraphics[width=0.6\textwidth]{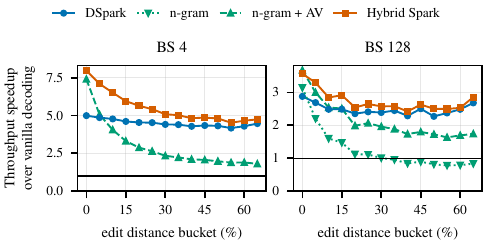}
  \vspace{-0.5em}
  \caption{The best drafter depends on content, and \sys tracks the better one.
  Output-throughput speedup over vanilla decoding on NextCoder (Qwen3-4B), by edit distance
  between the input program and the model's output (higher is better). The
  n-gram drafter outperforms \dspark only on near-copies, and verifying all of its
  drafts (n-gram) at BS~128 is slower than vanilla decoding from 35\% edit distance up.
  N-gram + AV gives it adaptive verification with our estimator.
  \Cref{tab:teaser} in the appendix lists the values.}
  \label{fig:teaser}
\end{figure}

Code editing makes this difference particularly clear. Given an
existing program and a request to modify it, the model often reproduces
large portions of the original code, changing only a few regions.
A model-based drafter can predict these tokens, but its draft length
is limited. An n-gram drafter can instead copy much longer
continuations directly from the prompt, at little cost. On the
NextCoder code-editing workload with Qwen3-4B at batch size
(BS)~4, \dspark, which drafts seven tokens per step, achieves
4.17--5.00\x speedup over vanilla decoding across edit distances
from 0\% to 70\%. An n-gram drafter that proposes up to 15 tokens achieves
7.42\x speedup when the output is nearly identical
to the input, but falls to 1.80\x on heavy rewrites
(\cref{fig:teaser}). Neither drafter is uniformly better:
copying wins when the output repeats the input, while model-based
drafting is more robust when the output changes substantially.

This creates a practical problem for serving systems. A deployment
serving both editing and general requests must either choose one
drafter for all traffic or maintain separate deployments for
different workloads. The former leaves performance on the table:
using \dspark everywhere misses the gains from copying on light
edits. The latter fragments GPU capacity across pools whose
traffic varies over time, requiring each to be provisioned for
its own peak. Ideally, a single deployment would use the better
drafter for each request, and adapt that choice as generation
progresses.

Combining drafters, however, is not as simple as running both and
taking the longer draft. The two drafters produce different kinds of
predictions, compete for the same verification budget, and must be
compared at every decoding step without introducing overhead.
This raises three challenges:
\begin{itemize}
    \item \textbf{(C1) Estimating acceptance for model-free drafts.}
    Model-based drafters provide token confidences, but an n-gram match
    only identifies a possible continuation, not how likely the target
    is to accept it. Without acceptance estimates, model-free drafts
    cannot participate in adaptive verification or be compared with
    model-based drafts.

    \item \textbf{(C2) Choosing between competing draft chains.}
    Each request now has two candidate chains that may agree, diverge,
    or differ entirely. Verifying both requires tree attention or
    increases verification cost, while choosing one requires a common
    scale and a price that reflects the rest of the batch.
    Existing hybrids~\citep{oliaro2024suffix} select the model-free
    chain when its score exceeds a fixed threshold, ignoring both
    the changing cost of verification and what other requests could
    gain from the same slots.

    \item \textbf{(C3) Making the decision practical at serving scale.}
    Drafter selection must run at every decoding step, entirely on
    the GPU and at negligible cost, without host-device
    synchronization. Otherwise, it undermines asynchronous scheduling
    and CUDA graph execution. Existing CPU-based hybrid
    approaches~\citep{oliaro2024suffix} incur this synchronization.
    The system must also preserve the target model's output
    distribution.
\end{itemize}

Our key observation is that adaptive verification already prices each
draft token: a verified slot costs the marginal goodput it could earn
elsewhere in the batch. If every proposed token carries an acceptance
estimate on a common scale, chains from different drafters can be
compared by their expected benefit at this price, which makes drafter
selection and budgeting one decision. Building on this observation,
\sys combines a model-based and a model-free drafter within one
adaptive-verification system. The model-free drafter receives
calibrated acceptance estimates from cheap features, fit once by
replaying target-model outputs on general text. At each step, \sys
selects a chain for each request at the common price and allocates
the batch's budget across the selected chains, all on the GPU and
without changing the output distribution.

We instantiate \sys with \dspark and a GPU n-gram drafter in
vLLM~\citep{kwon2023vllm}, and evaluate it across five models,
seven workloads, and three batch sizes. Across five code-editing
workloads, \sys achieves 1.10--1.62\x the output throughput
of \dspark at BS~4 and 1.04--1.27\x at BS~128, while
remaining at 0.97--1.01\x of \dspark on chat at $T{=}0$.
The gains follow how much of the output can be reused:
on NextCoder, light edits (15--20\% edit distance) achieve
1.28--1.33\x at BS~4, while heavy edits (60--70\%)
achieve 1.03--1.12\x (\cref{fig:main}). A single deployment
thus gains from copying when the output repeats its input
and remains robust on general text.

The same split between learned prediction and reuse appears in
reinforcement learning (RL) post-training. Rollout generation dominates
training time, and generations from the same prompt often overlap
across adjacent epochs~\citep{liu2025specrl,he2025rhymerl}. As short
rollouts finish, a few long generations at small effective batch size
determine step time~\citep{shao2026das}, which is where cheap copying
helps most. History-based drafters exploit this
overlap~\citep{he2025rhymerl,shao2026das}, while a model-based drafter
can go stale as the policy changes~\citep{shao2026das}. \sys's
model-free slot can hold such a drafter, and we leave RL rollout
generation to future work.

Our contributions are:
(1)~a characterization of how drafter choice depends on
content and batch size (\cref{sec:motivation});
(2)~a recipe for calibrating model-free drafts using offline
replay and a per-position logistic estimator, instantiated
for n-gram drafting and shown to transfer across datasets
and models (\cref{sec:estimator,sec:eval-components});
(3)~a drafter-agnostic chain-selection rule based on the
marginal goodput of verification slots, integrated with
adaptive verification entirely on the GPU, including an
analysis of stale budget sizing
(\cref{sec:selection,sec:budget,sec:eval-stale});
and (4)~lossless alternative policies for probabilistic
drafting and an evaluation of the \dspark + n-gram
instantiation across models, workloads, batch sizes,
and temperatures
(\cref{sec:lossless,app:lossless,sec:eval}).

\section{Background}
\label{sec:background}

\paragraph{Speculative decoding and goodput.}
In each step, a \emph{drafter} proposes a \emph{draft chain} of up to $K$
tokens per request. The target model scores all of them in one forward pass and
accepts the longest prefix that passes rejection
sampling~\citep{leviathan2023fast,chen2023accelerating}, plus one token of its
own. The \emph{acceptance length} (AL) is the mean number of tokens emitted per
request per step. We write $\alpha_i$ for the conditional acceptance rate of
draft position $i$ given that positions $1,\dots,i-1$ were accepted, and $S_k =
\prod_{i \le k}\alpha_i$ for the \emph{survival} of position $k$, the
probability that it is accepted. \emph{Goodput} is the expected number of tokens
emitted per unit of step time. Output throughput, our end-to-end metric, is
total generated tokens over the duration of a benchmark run.

\paragraph{Drafters.}
Model-based drafters run a neural network, a small model or heads on the
target (\cref{sec:related}). \dspark~\citep{cheng2026dspark} drafts a
$K{=}7$-token block semi-autoregressively and has a confidence head that
predicts each draft token's conditional acceptance. Model-free drafters
(\cref{sec:related}) copy continuations from the context or a datastore, such
as earlier rollouts of the same prompt in RL
post-training~\citep{he2025rhymerl,shao2026das}. They
cost almost nothing to run and can draft long chains when the output copies its
input, but they give no calibrated confidence. Our instantiation uses the simplest of
them, an n-gram match against the request's own tokens.

\paragraph{Adaptive verification (AV).}
Verifying one more draft token, a \emph{slot}, is nearly free at small batch size
but costly at large batch size, once the verification pass becomes compute
bound~\citep{liu2026illusion}. With long contexts, loading the KV cache keeps
large batches memory bound longer, so SD can still
help there~\citep{sadhukhan2025magicdec}. \dspark's adaptive verification (\av) treats all
slots in the batch as one \emph{verification budget} $b$. It models the
step cost $C(b)$ from profiled tables and the expected emitted tokens
$A(b) = B + \sum_{\text{chosen slots}} S$ for $B$ requests, and chooses the
budget $b^*$ that maximizes goodput $G(b) = A(b)/C(b)$. It gives the $b^*$
slots to the (request, position) pairs with the highest survival, so each request keeps
a prefix of its chain.

\paragraph{Asynchronous scheduling.}
Modern engines overlap the host's preparation of step $t{+}1$ with the GPU's
execution of step $t$. The host must fix the number of tokens in the next
verification pass before the GPU has produced the confidences for it. \dspark
therefore sizes the budget on the host from confidences produced two steps
prior, one step older than the ones the GPU uses to allocate
it~\citep[\S5.2]{cheng2026dspark}. Any per-step
decision that \sys adds must respect this split: a host read of a GPU value
mid-step would serialize the pipeline.

\section{Motivation}
\label{sec:motivation}

We measure the two drafters on Qwen3-4B with NextCoder code
edits~\citep{aggarwal2025nextcoder}, bucketed by normalized edit distance
between the input program and the model's output, and on the general-text
SPEED-Bench~\citep{abramovich2026speedbench}. \Cref{sec:eval-setup} gives details.

\obs{1: The better drafter depends on the request's content}
\Cref{fig:teaser} shows that with \av, the n-gram drafter outperforms \dspark at BS~4 only
below 10\% edit distance, and above that \dspark's speedup is up to 2.5 times the
n-gram drafter's. Within a
request, the drafter that accepts more also alternates from step to step:
\citet{liu2026illusion} show this for an n-gram drafter and EAGLE, and an oracle that picks the
better drafter per step gains beyond either alone~\citep[Figs.~11
and~12]{liu2026illusion}. This motivates choosing a drafter per request and per
step (\cref{sec:selection}).

\begin{figure}[t]
  \centering
  \includegraphics[width=0.6\textwidth]{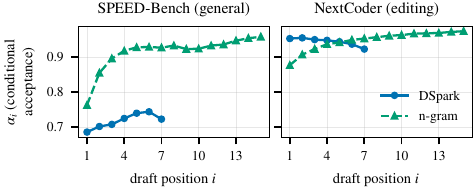}
  \vspace{-0.5em}
  \caption{The two drafters have opposite acceptance profiles. Conditional
  acceptance $\alpha_i$ by draft position (Qwen3-4B; higher is better). Given a
  match, the n-gram drafter's first token is its weakest and its deep tokens are
  accepted more often than any of \dspark's. The n-gram curve counts only steps
  with a match.}
  \label{fig:alpha}
\end{figure}

\obs{2: N-gram drafts are weakest at the first token and strongest deep in the chain}
SD analyses often summarize a drafter by one acceptance rate~\citep{leviathan2023fast},
but the n-gram drafter does not fit this summary (\cref{fig:alpha}). Its first
token is its least reliable, because a match shows only that the context
repeats, not that the output will copy it. Given a match, this token is accepted
at 0.88 on NextCoder (0.76 on SPEED-Bench). Once it is accepted, the output has
started to copy the context and tends to continue: acceptance rises to 0.97 by
the fifteenth token (0.96), and positions 8--15, which \dspark never drafts, are
accepted at 0.96--0.97 (0.92--0.96). \dspark's acceptance is flatter, at
0.92--0.96 (0.69--0.74). Because
\av ranks individual slots, these per-position differences determine which slots it allocates.
The estimator therefore predicts acceptance per position, with separate terms for
the first token (\cref{sec:estimator}). When the n-gram chain begins with
\dspark's tokens, \sys keeps its tail beyond \dspark's seventh token, the
most reliable tokens in either chain (\cref{sec:selection}).

\obs{3: The n-gram drafter needs the budget at large batch size but loses throughput to it at small batch size}
Verifying every n-gram draft token is cheap at BS~4, but at BS~128 it is slower
than vanilla decoding on every NextCoder bucket from 35\% edit distance up, down to 0.78\x
(\cref{fig:teaser}). Giving the same drafter \av with our estimator
(\cref{sec:estimator}) yields a 1.90\x output-throughput speedup over verifying
every draft on pooled NextCoder at BS~128, from 1.17\x at 0--5\% to 2.15\x at
60--65\%. At BS~4, where verification is nearly free, the budget has little to
gain, and the same \av costs the n-gram drafter 4.2\% throughput. One cause is
that the budget is sized from estimates produced two steps prior
(\cref{sec:background}). Because n-gram matches appear and vanish from step to
step, these estimates are often wrong. Against a budget sized from fresh
survivals, this loses 7--9\% of predicted goodput at BS~4
(\cref{sec:eval-stale,app:stale}). The n-gram drafter must therefore
join the shared budget with calibrated confidences
(\cref{sec:estimator,sec:budget}), and \sys must keep the cost of stale sizing
small at small batch size (\cref{sec:eval-stale}).

\section{Design}
\label{sec:design}

\begin{figure*}[t]
\centering
\resizebox{\linewidth}{!}{%
\begin{tikzpicture}[
  font=\footnotesize,
  box/.style={draw, rounded corners=2pt, align=center, minimum height=2.1em, inner sep=3pt, fill=white},
  gpu/.style={box, fill=blue!6},
  host/.style={box, fill=orange!10},
  arr/.style={-{Stealth[length=4pt]}, thick},
  darr/.style={-{Stealth[length=4pt]}, thick, dashed},
  ]
  \node[gpu] (tgt) at (0,0) {Target model\\verify step $t$};
  \node[gpu] (ds) at (3.1,0.65) {\dspark drafter\\chain $d$, confidences $c_k$};
  \node[gpu] (ng) at (2.5,-0.65) {n-gram scan\\chain $g$; $n$, $a$, $m$};
  \node[gpu] (est) at (5.0,-0.65) {Estimator\\$h_k = \sigma(\theta^\top\phi_k)$};
  \node[gpu, minimum width=3.1cm] (sel) at (8.4,0) {Chain selection\\$V = \sum_k \max(S_k-\lambda,0)$\\+ contains rule};
  \node[gpu] (alloc) at (11.6,0) {\av allocation\\top-$b^*$ slots\\by fresh $S$};
  \node[gpu] (ver) at (14.2,0) {Target model\\verify step $t{+}1$};
  \node[host] (size) at (11.6,-1.85) {\av budget sizing\\$b^*$ and $\lambda = G(b^*)\,p$\\from profiled cost tables};
  \draw[arr] (tgt.north) |- (ds.west);
  \draw[arr] (tgt.south) |- (ng.west);
  \draw[arr] (ng) -- (est);
  \coordinate (inx) at ($(sel.west)+(-0.25,0)$);
  \draw[arr] (ds.east) -| ($(inx)+(0,0.2)$) -- ($(sel.west)+(0,0.2)$);
  \draw[arr] (est.east) -| ($(inx)+(0,-0.2)$) -- ($(sel.west)+(0,-0.2)$);
  \draw[arr] (sel) -- node[above, font=\scriptsize]{$S$} (alloc);
  \draw[arr] (alloc) -- (ver);
  \draw[darr] ($(sel.south)+(-0.5,0)$) |- node[pos=0.72, below, font=\scriptsize]{stale $S$ (async copy)} (size.west);
  \draw[darr] (size.north) -- node[right, font=\scriptsize]{$b^*$} (alloc.south);
  \draw[darr] ($(size.north west)+(0.25,0)$) -- node[midway, right=2pt, font=\scriptsize]{$\lambda$} ($(sel.south)+(0.7,0)$);
  \node[font=\scriptsize, text=blue!60!black, anchor=west] at (-0.6,1.25) {GPU, every step};
  \node[font=\scriptsize, text=orange!60!black, anchor=west] at ($(size.east)+(0.1,0)$) {CPU host, one step behind};
\end{tikzpicture}}
\vspace{-0.5em}
\caption{\sys overview, instantiated with \dspark and an n-gram drafter. Every
per-step decision runs on the GPU (blue), while the CPU host (orange) sizes the
budget and sets the slot price $\lambda$ from survivals produced two steps
prior. Solid
arrows stay on the GPU, and dashed arrows carry host scalars or asynchronous
copies.}
\label{fig:overview}
\end{figure*}
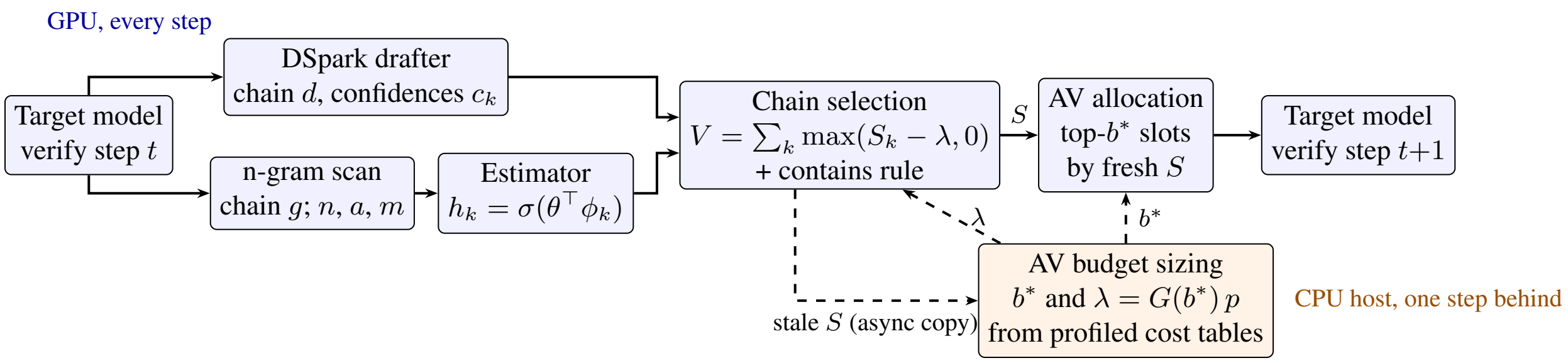

\subsection{Overview}
\label{sec:overview}

\sys addresses the three challenges through three design choices (\cref{fig:overview}). (C1) It estimates acceptance probabilities for the model-free drafter, putting both chains on a common scale of prefix survival probabilities (\cref{sec:estimator}). (C2) It selects one chain per request by comparing their expected value against the opportunity cost of verification slots elsewhere in the batch (\cref{sec:selection}). (C3) It makes all token-dependent decisions on the GPU, using budget parameters from the previous step to avoid host-device synchronization while preserving lossless decoding (\cref{sec:budget}).

\textbf{Interfaces.} \sys requires each drafter to provide a candidate chain and per-position acceptance estimates. The model-based drafter returns a chain $d$ with calibrated confidences $c_k$, while the model-free drafter returns a chain $g$ and features from which an estimator predicts acceptance probabilities $h_k$. The model-free chain is fixed before verification and independent of the model-based drafter's sampled tokens. Both drafters run on the GPU without a host read. Selection, budget allocation, and lossless policies depend only on $d$, $g$, $c_k$, and $h_k$, so they do not change when either drafter is replaced.

\textbf{One step.} \Cref{alg:main} illustrates one \sys step. \dspark proposes $K{=}7$ tokens with confidences $c_k$, while the n-gram drafter proposes $L{=}15$ tokens following the longest earlier occurrence of the request's last 5--15 tokens (line~\ref{ln:scan}). The estimator predicts acceptance probabilities $h_k$ (line~\ref{ln:est}), and both chains' prefix survivals are computed as running products of their estimates. \sys selects the chain with the larger net value at slot price $\lambda$ (lines~\ref{ln:contains}--\ref{ln:value}), and \av allocates the $b^*$ verification slots across the selected chains by survival (line~\ref{ln:alloc}). All decisions run on the GPU, and the host computes $b^*$ and $\lambda$ from survivals produced two steps prior (line~\ref{ln:copy}), so no mid-step synchronization is needed.

Beyond the evaluated \dspark and n-gram pair, the interface admits datastore-based drafters, such as the suffix trees of SuffixDecoding~\citep{oliaro2024suffix} or suffix trees over earlier RL rollouts~\citep{shao2026das}, and more than two drafters, which we leave to future work. \Cref{tab:notation} in the appendix summarizes the notation.

\begin{algorithm}[t]
\caption{One \sys drafting step with \dspark and an n-gram drafter.}
\label{alg:main}
\begin{algorithmic}[1]
\REQUIRE context $x_r$, \dspark chain $d_r$ with confidences $c_{r,1..K}$,
coefficients $\theta$; host scalars $\lambda$ and $b^*$
\FORALL{requests $r$ in parallel}
  \STATE $(g_r, n, a, m) \gets \textsc{NgramScan}(x_r)$ \lnlabel{ln:scan}
  \STATE $h_k \gets \sigma(\theta^\top \phi_k(n,a,m))$, \; $S^g_k \gets \prod_{j\le k} h_j$ \lnlabel{ln:est}
  \STATE $S^d_k \gets \prod_{j \le k} c_{r,j}$
  \IF{$|g_r| > K$ \AND $g_{r,1..K} = d_r$} \lnlabel{ln:contains}
    \STATE keep $g_r$; $S_k \gets \max(S^g_k, S^d_k)$ for $k \le K$, then
    continue with $h_k$ \lnlabel{ln:merge}
  \ELSIF{$g_r$ matched \AND $\sum_k (S^g_k - \lambda)^+ \ge \sum_k (S^d_k - \lambda)^+$} \lnlabel{ln:value}
    \STATE keep $g_r$ with $S \gets S^g$
  \ELSE
    \STATE keep $d_r$ with $S \gets S^d$
  \ENDIF
\ENDFOR
\STATE \av: give the top-$b^*$ (request, position) slots by $S$; each request
keeps a prefix of its chain \lnlabel{ln:alloc}
\STATE copy $S$ to the host asynchronously for the next budget sizing \lnlabel{ln:copy}
\end{algorithmic}
\end{algorithm}

\subsection{Calibrated acceptance estimates for n-gram drafts}
\label{sec:estimator}

\begin{figure}[t]
  \centering
  \includegraphics[width=0.6\textwidth]{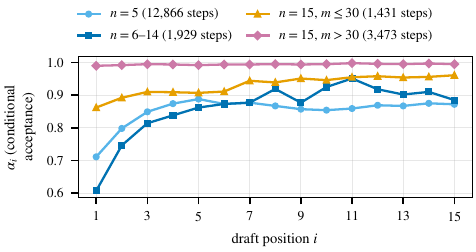}
  \vspace{-0.5em}
  \caption{Full matches and long copies raise acceptance most on the first
  token. Conditional acceptance
  $\alpha_i$ by draft position, grouped by match length $n$ and, for full
  matches, copy length $m$ (Qwen3-4B, the half of SPEED-Bench the estimator is
  fit on; higher is better).}
  \label{fig:alpha-by-match}
\end{figure}

\textbf{Challenge.} Unlike model-based drafters, n-gram drafters produce tokens without confidence scores (C1). \av ranks slots across requests and drafters by survival and sizes the budget from it, so an n-gram draft needs a calibrated survival estimate to take part. We obtain these estimates from features already available during n-gram matching.

\textbf{Mechanism.} We predict the conditional acceptance probability $h_k$ at each draft position $k$ using a logistic regression:
\begin{equation}
h_k = \sigma\big(\beta_k + w_{\text{pos}(k)}\,(n-5) + \delta\, a
 + [n{=}15]\,(\ell_{\text{pos}(k)} + \kappa_{\text{pos}(k)} \ln\tfrac{\min(m,256)}{15})\big),
\label{eq:estimator}
\end{equation}
where $\sigma$ is the logistic function, $\text{pos}(k)$ distinguishes the first token from later tokens, and $S^g_k = \prod_{j \le k} h_j$ gives the prefix survival probability.

The estimator has one intercept $\beta_k$ per draft position, which captures the rising acceptance profile of Observation~2. It also uses four match features:
\begin{itemize}
    \item \textbf{Match length} $n \in [5,15]$.
    \item \textbf{Agreement} $a$, which is 1 if the runner-up match predicts the same next token or there is no runner-up. Agreement matters because two matches that continue differently leave open which one the output copies.
    \item \textbf{Full-match bonus} $\ell$, applied when the match reaches the 15-token cap.
    \item \textbf{Copy length} $m$, how far a full match extends backwards, capped at 256 tokens to bound the search.
\end{itemize}
Match length, the full-match bonus and copy length each have separate first-token and later-token coefficients ($w$, $\ell$ and $\kappa$, indexed by $\text{pos}(k)$).

As \cref{fig:alpha-by-match} shows, full matches and long copies improve acceptance most strongly at the first position. When a full match extends beyond 30 tokens, acceptance exceeds 99\% at every position. For shorter copies, first-token acceptance is 86\%, rising to 89--96\% at later positions. The scan already computes $n$, $a$, and $m$, so the estimate adds only one fused elementwise step per request.

\textbf{Fitting once on general data.}\label{sec:fitting} We fit the estimator once per target model using offline acceptance labels from greedy decoding. Since the target output is deterministic at $T{=}0$, replaying each generated sequence and rerunning the n-gram scan on its prefixes gives exact acceptance labels without online verification. This approach applies to any model-free drafter whose proposals can be reconstructed from the context and datastore.

We generate up to 512 tokens for 1,530 prompts from SPEED-Bench's throughput split~\citep{abramovich2026speedbench}, using an 8k-token input limit. Each of the three entropy categories is split approximately in half, yielding 764 fitting prompts and 766 held-out prompts. We fit the regression using Newton's method with ridge regularization and select the logarithmic copy-length feature using held-out log-loss. None of the editing benchmarks (Blazedit, CanItEdit, NextCoder, EditBench) are used for fitting or feature selection, so all editing results are out of sample. \Cref{app:coef} reports the coefficients and fitting data.

\subsection{Choosing one chain per request under one budget}
\label{sec:selection}

\textbf{Challenge.} Each request has two candidate chains, but verifying both requires tree attention or extra slots (C2). Choosing one must account for the rest of the batch: a slot spent on one request cannot be used by another. Two simple policies fall short (\cref{sec:eval-policies}): always preferring an n-gram match discards \dspark's chain even for unreliable matches, while choosing the chain with more expected accepted tokens, $\sum_k S_k$, favors long chains with low-value tails.

\textbf{Slot price.} We assign each verification slot a price based on its
opportunity cost elsewhere in the batch. \av selects the budget $b^*$ that
maximizes goodput $G(b) = A(b)/C(b)$, where $A(b)$ is the expected number of
emitted tokens and $C(b) = C_0 + p\,b$ is the step cost with slot cost $p$
(\cref{app:cost}). Adding a slot with survival $s$ changes goodput by
\begin{equation}
\Delta G = \frac{A + s}{C + p} - \frac{A}{C} \approx \frac{s - G\,p}{C}
= \frac{s - \lambda}{C},
\label{eq:price}
\end{equation}
where $G=A/C$. Thus, a slot improves goodput only when $s>\lambda$, with $\lambda=G(b^*)p$. We use this single slot price across the batch. The price rises with batch size (\cref{app:value}), so chain selection matters most at BS~128.

\textbf{Chain value.} We score each chain by the expected benefit of its slots after subtracting their price:
\begin{equation}
V = \textstyle\sum_k \max(S_k-\lambda,0).
\label{eq:value}
\end{equation}
\sys selects the chain with larger $V$, breaking ties in favor of the n-gram chain (line~\ref{ln:value}). Since survival decreases along a chain, the slots with $S_k>\lambda$ form a prefix, matching the slots \av would verify. Unlike $\sum_k S_k$, this value ignores low-survival tail slots, so a longer chain wins only when its extra slots justify their cost (\cref{app:value}).

\textbf{Contains rule.} When the n-gram chain is longer and begins with all $K$ tokens from \dspark, it is accepted at least as far as \dspark's chain on every target draw. We therefore keep the n-gram chain and merge their survival estimates: the first $K$ positions take the larger survival, and the rest use the n-gram estimates (lines~\ref{ln:contains} and~\ref{ln:merge}). This combines \dspark's stronger estimates at the start with the n-gram tail, where the n-gram drafter is strongest (Observation~2).

\subsection{Sharing the verification budget}
\label{sec:budget}

\textbf{Challenge.} \av assumes fixed-length drafts from one drafter. With \sys, chain lengths and survival curves vary across requests and steps, while selection and allocation must remain on the GPU without host synchronization (C3).

\textbf{Mechanism.} \sys keeps \av's budget sizing and allocation rules unchanged. The selection kernel writes the chosen chains, their valid lengths, and merged survivals into fixed-size GPU buffers. \av allocates verification slots and constructs the verification layout directly on the GPU, supporting variable-length chains without a host round trip.

The host computes the next budget $b^*$ and slot price $\lambda=G(b^*)\,p$ from survivals produced two steps prior, one step older than the ones the GPU allocates from (line~\ref{ln:copy}). Both are passed to the GPU as scalar launch parameters, avoiding synchronization within the current step. \Cref{app:cost} describes the cost model.

\subsection{Losslessness}
\label{sec:lossless}

\sys is lossless at any target sampling temperature when drafting is greedy, but its value-based selection is not generally lossless with probabilistic drafting. With greedy drafting, both chains and all selection and allocation decisions depend only on the context, and the target verifies each draft token as a point mass. The same argument applies to D-cut pruning~\citep[App.~A]{liu2026dcut}.

With probabilistic drafting, the contains and value rules depend on \dspark's sampled tokens. Whether a slot is verified can then depend on the token sampled there, biasing the output distribution~\citep[App.~B]{liu2026dcut}.

Two simpler policies remain lossless: \emph{n-gram-first}, which selects the n-gram chain whenever a match exists, and \emph{extension}, which verifies \dspark's chain unchanged and appends the n-gram tail only when the entire \dspark chain is its prefix. \Cref{app:lossless} provides the arguments and summarizes the guarantees in \cref{tab:lossless}.

\section{Evaluation}
\label{sec:eval}

\begin{figure*}[t]
  \centering
  \includegraphics[width=\textwidth]{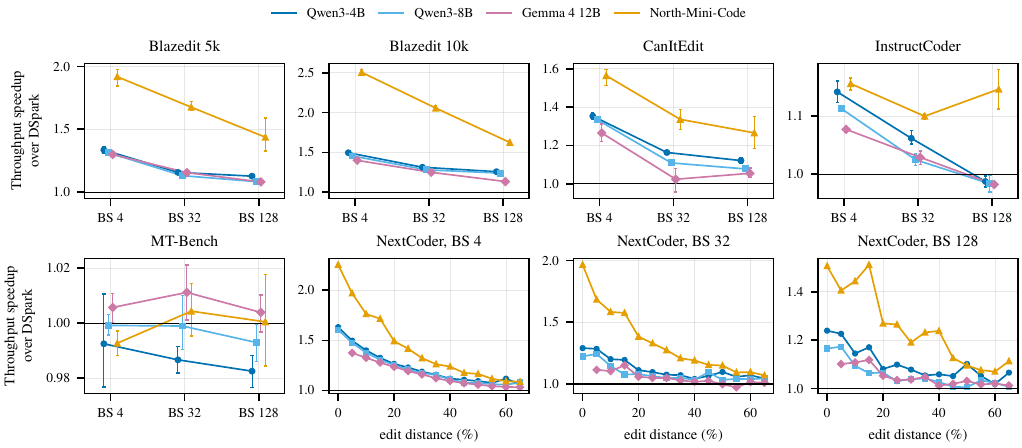}
  \vspace{-1.5em}
  \caption{\sys raises output throughput over \dspark on almost every editing
  cell, and the gain shrinks with batch size and with edit distance. On chat
  (MT-Bench), it stays within 0.97--1.01\x. Speedup in output throughput
  (higher is better), mean of 3 repeats. First five panels: by batch size, with
  95\% CIs. Last three panels:
  NextCoder per edit-distance bucket. Missing points are buckets with fewer
  than 10 samples.}
  \label{fig:main}
\end{figure*}

We evaluate \sys along seven questions:
(Q1) How much does it improve editing, and how does the gain change with batch size and edit distance?
(Q2) Does it hurt performance on other workloads?
(Q3) Would a simpler policy perform as well?
(Q4) Which components matter, and does the estimator transfer across settings?
(Q5) What is the runtime overhead of the additional drafter?
(Q6) How much goodput is lost by sizing the budget from stale survival estimates?
(Q7) How does it behave at T=1?

\Cref{sec:eval-ood} separately examines models and contexts where \dspark's drafter is weak.

\subsection{Setup}
\label{sec:eval-setup}

\paragraph{Implementation, hardware and serving.}
We evaluate the instantiation of \cref{sec:design} (\dspark and the n-gram
drafter), implemented in vLLM and referred to as \sys below. Every arm runs on vLLM's V1 engine with
Model Runner V2 (MRv2). All runs use NVIDIA H100 80\,GB (HBM3) GPUs. BS~4 and
32 run on one GPU and BS~128 on two or four GPUs with
tensor parallelism, with the same layout for every arm of a comparison.
All runs enable asynchronous scheduling and disable prefix caching. Every arm that uses \av enables it at every batch size.

\paragraph{Models and drafters.}
We evaluate five target models, each with its own \dspark drafter ($K{=}7$):
Qwen3-4B and Qwen3-8B~\citep{yang2025qwen3}, Gemma~4
12B~\citep{gemma2026gemma4}, North-Mini-Code-1.0~\citep{cohere2026northminicode},
a 30B-total, 3B-active mixture-of-experts code model served with 4-bit weights,
and Llama-3.1-8B-Instruct~\citep{grattafiori2024llama3}. \Cref{tab:models}
lists the checkpoints. The n-gram drafter matches 5--15 tokens
(longest match wins) and drafts up to 15. Each model uses its own estimator
coefficients, fit as in \cref{sec:fitting}.

\paragraph{Workloads.}
\Cref{tab:workloads} in \cref{app:workloads} lists their sizes, prompts per
run and output limits. For NextCoder and
EditBench~\citep{chi2026editbench}, we bucket requests by the normalized
Levenshtein distance between the input program and the longest code block of
the target model's own greedy output without SD, in 5\% buckets. We keep
inputs above 1{,}000 characters and buckets with at least 10 samples. Buckets
are therefore model specific. EditBench requests with more than 6{,}000 tokens
of context (prompt plus output) are out of distribution for the \dspark drafters
and are excluded from the main results (\cref{sec:eval-ood}). Blazedit 5k and
10k~\citep{blazedit2025} and CanItEdit~\citep{cassano2023canitedit} keep inputs
above 1{,}000 characters with reference edit distance 5--25\%. We use
InstructCoder~\citep{li2023instructcoder} as a mixed workload and
MT-Bench~\citep{zheng2023mtbench} first turns as general chat.

\paragraph{Metric and statistics.}
We use \texttt{vllm bench serve} with the chat-completions API, a temperature of 0 unless otherwise stated, and client concurrency equal to the batch size (BS). Output throughput is the total number of output tokens divided by the run duration. Speedup is \sys's output throughput divided by the baseline's, so values above 1 indicate that \sys is faster. For bucketed workloads, we pool throughput across buckets: total output tokens divided by the total duration, treating the buckets as if they ran back to back. Each cell reports the mean over 3 independent runs (5 for the ablation), each launched on a fresh server. We report 95\% confidence intervals computed by bootstrap resampling over the repeats. If a speedup's confidence interval includes 1, we call the two systems statistically indistinguishable, or \emph{level}.

\paragraph{Baselines.}
(1)~Vanilla decoding. (2)~\dspark with its own \av and the released block size $K{=}7$, the primary
baseline. (3)~The n-gram drafter with \av,
using \sys's estimator for its confidences (n-gram + \av). (4)~The n-gram drafter
with every draft verified. (5)~Simpler hybrid policies and ablations in
\cref{sec:eval-policies,sec:eval-components}.

\subsection{Code editing (Q1)}
\label{sec:eval-editing}

\textbf{\sys's gain is largest at small batch size and shrinks at large batch
size.} Across the five editing workloads and the three in-distribution models,
\sys's speedup over \dspark is 1.10--1.62\x at BS~4 and 1.04--1.27\x
at BS~128 (\cref{fig:main}; \cref{tab:main-all} in the appendix lists every
cell). Every one of these speedups has a CI that excludes 1, and the only level
editing cell is Gemma~4 12B on CanItEdit at BS~32. NextCoder gains least
because it includes buckets up to 70\% edit distance, where little of the
output is copied.

\sys's gains are largest where \dspark's drafter is weak: on North-Mini-Code
(1.28--2.50\x at BS~4), on Llama-3.1-8B, and on EditBench requests with
more than 6{,}000 context tokens (up to 7.58\x). There \dspark's drafter accepts
few tokens per step, which leaves more room for the n-gram drafter. A
better-trained drafter would likely reduce these gains, so we report them
separately in \cref{sec:eval-ood}.

\textbf{The gain comes from longer accepted chains.} On Blazedit and CanItEdit
at BS~4, \dspark accepts 6.1--6.9 tokens per step for the three models,
and \sys 8.4--10.0 (\cref{fig:al} in the appendix shows AL per model and
workload). The gain shrinks with batch size because $\lambda$
grows, so fewer n-gram tail slots are admitted to the budget.

\textbf{The gain tracks how much the output copies.} The last three panels of
\cref{fig:main} break NextCoder down by edit distance: as less of the output is
copied from the input, the gain falls. At BS~4 it falls from 1.61--1.63\x
for the two Qwen3 models at 0--5\% to 1.03--1.12\x at 60--70\%, and at
BS~128 to 1.01--1.07\x. EditBench shows the same trend
(\cref{fig:edit-eb} in the appendix).

\subsection{Workloads where the n-gram drafter rarely helps (Q2)}
\label{sec:eval-nonedit}

\textbf{\sys is 0.97--1.01\x of \dspark on chat.} On MT-Bench at T{=}0 with
greedy drafting, this range covers every model and batch size
(\cref{fig:main,tab:policy}). AL is
the same for both arms to within 0.1 (\cref{fig:al}) because the n-gram drafter rarely finds a match worth its
price. On InstructCoder, where outputs partly copy short inputs, \sys achieves
1.08--1.14\x at BS~4 and 1.02--1.06\x at BS~32, and is 0.98--0.99\x at BS~128
(Gemma~4 12B: 0.98\x [0.98, 0.99]).

\subsection{Simpler policies (Q3)}
\label{sec:eval-policies}

\textbf{The n-gram drafter alone is not a substitute.} With \av, the n-gram
drafter reaches 2.70\x over vanilla decoding on pooled NextCoder at BS~4 and 2.04\x at
BS~128 (Qwen3-4B, \cref{fig:teaser}). \sys reaches 5.52\x and 2.75\x, a
speedup of 2.04\x [2.02, 2.07] and 1.34\x [1.33, 1.36] over the n-gram drafter.

\textbf{Always taking a matching n-gram chain gives up much of the gain at large
batch size.} \Cref{tab:policy} compares the proposed policy with
n-gram-first and the extension policy on two NextCoder buckets and MT-Bench. At T{=}0,
n-gram-first's throughput is 2.5--5.3\% lower than the proposed policy's at
BS~4 and 5.1--7.3\% lower at BS~128, and on 40--45\% at BS~128 it is slower than \dspark alone
(0.98\x). Its AL there is 6.5, against 7.2--7.4 for the proposed and extension
policies and 6.1 for \dspark: it discards \dspark's chain on short,
unreliable matches.

\begin{table}[t]
\centering
\small
\caption{Output-throughput speedup of each policy over \dspark with greedy
drafting at the same temperature (Qwen3-4B, 3 repeats; $^\circ$: CI includes
1). T{=}1 shows each policy where it is lossless: the proposed policy with
greedy drafting, the others with probabilistic drafting.}
\label{tab:policy}
\setlength{\tabcolsep}{3pt}
\footnotesize
\begin{tabular}{llrrrrrr}
\toprule
 & & \multicolumn{4}{c}{NextCoder} & \multicolumn{2}{c}{MT-Bench} \\
\cmidrule(lr){3-6}
 & & \multicolumn{2}{c}{15--20\%} & \multicolumn{2}{c}{40--45\%} & & \\
\cmidrule(lr){3-4}\cmidrule(lr){5-6}\cmidrule(lr){7-8}
BS & Policy & T{=}0 & T{=}1 & T{=}0 & T{=}1 & T{=}0 & T{=}1 \\
\midrule
\multirow{4}{*}{4}
  & Proposed       & 1.30 & 1.28 & 1.13 & 1.13 & 0.99 & 0.98 \\
  & N-gram-first   & 1.27 & 1.23 & 1.07 & 1.06 & 0.98 & 1.01 \\
  & Extension      & 1.27 & 1.22 & 1.13 & 1.10 & 1.00$^\circ$ & 1.02 \\
  & \dspark (prob.) &     & 0.98 &      & 0.97 &      & 1.03 \\
\midrule
\multirow{4}{*}{128}
  & Proposed       & 1.16 & 1.18 & 1.05 & 1.07 & 0.97 & 0.97 \\
  & N-gram-first   & 1.10 & 1.09 & 0.98 & 0.96$^\circ$ & 0.97 & 0.96 \\
  & Extension      & 1.11 & 1.07 & 1.06 & 1.02$^\circ$ & 0.95 & 0.98$^\circ$ \\
  & \dspark (prob.) &     & 0.96$^\circ$ & & 0.98$^\circ$ & & 0.99$^\circ$ \\
\bottomrule
\end{tabular}
\end{table}

\textbf{Both the slot price and the contains rule improve throughput.} In the ablation in
\cref{sec:eval-components}, replacing the value rule with the expected
accepted length, $\sum_k S_k$, reduces throughput by 1.0\%, and removing the contains
rule costs 2.4\%, the largest single effect (\cref{fig:ablation}). Without the
contains rule, AL falls from 7.4 to 6.9, and the share of requests on the n-gram
chain falls from 26.5\% to 15.5\%.

\subsection{Estimator features and transfer (Q4)}
\label{sec:eval-components}

\begin{figure}[t]
  \centering
  \includegraphics[width=0.6\textwidth]{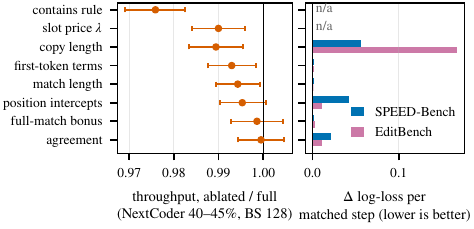}
  \vspace{-0.5em}
  \caption{The contains rule and the slot price matter most for throughput, and
  copy length matters most for calibration. Left: output throughput with one
  component removed, relative to full \sys (Qwen3-4B, NextCoder 40--45\%,
  BS~128, 5 repeats, 95\% CIs; higher is better). Right: increase in log-loss of
  the estimator per matched step when the feature is removed (lower is
  better). The contains rule and the slot price are selection rules, not
  estimator features, so calibration does not apply to them (n/a).}
  \label{fig:ablation}
\end{figure}

\textbf{No single estimator feature matters much for throughput, but together
they do.} We remove one component
at a time from \sys (Qwen3-4B). On NextCoder 40--45\% at BS~128
(\cref{fig:ablation}, left), besides the two selection rules above, only copy
length, the first-token terms and match length have a measurable effect
(0.6--1.1\%). In three more cells, every ablation stays within 0.992--1.003\x
of full \sys. A small leave-one-out cost does not make a feature redundant:
the match features overlap, so the refit shifts weight onto the rest. The
agreement term shows no leave-one-out effect, yet the first token is accepted
24\% of the time when the two matches disagree, against 81\% otherwise.
Removing all match features at once from the n-gram drafter with \av costs it
4.1\% (0.959\x [0.951, 0.967]) on pooled NextCoder at BS~128, whereas removing
any one has no measurable effect.
Adding terms one at a time (\cref{app:ladder}) shows that
most of the gain comes from giving the n-gram drafter any calibrated survival:
with one shared intercept and no match features, it reaches 0.973\x of the
full estimator's throughput on two NextCoder buckets at BS~128. Copy length
closes most of the rest. AL is unchanged, so the estimator helps by choosing
which slots to verify.

Calibration differs (\cref{fig:ablation}, right). Copy length
dominates on EditBench, where long copies are common: without it, the expected
calibration error (ECE)~\citep{guo2017calibration} rises from 0.032 to 0.132.
On held-out SPEED-Bench, the position intercepts matter most (ECE 0.012 to
0.029).

\textbf{The estimator transfers across datasets and models.} The estimator is
calibrated only on SPEED-Bench, a general-text benchmark, so all editing
results are out of sample. Even so, its calibration error on EditBench is only 0.032. It also transfers across models. Serving Qwen3-4B with coefficients fitted on
Qwen3-8B or Llama-3.1-8B changes pooled NextCoder throughput by at most 0.7\%
at any batch size (0.996--1.007\x relative to model-specific coefficients), and
AL by at most 0.08 in every bucket.

\subsection{Overhead (Q5)}
\label{sec:eval-overhead}

\textbf{Running the n-gram drafter alongside \dspark costs at most 1.2\%.} To
isolate this overhead, we configure \sys never to select the n-gram chain, so it
performs the scan, estimation and selection but verifies exactly \dspark's
chains. Its pooled
throughput relative to \dspark at BS~128 is 0.988\x on both NextCoder and
EditBench for Qwen3-4B, and 0.993\x and 0.996\x for Qwen3-8B.

\subsection{Stale budget sizing (Q6)}
\label{sec:eval-stale}

\textbf{Sizing the budget from stale survivals costs at most 0.64\% of goodput
at BS~128.} We compare the goodput of the budget sized from stale survivals
with that of a budget sized from fresh ones (\cref{app:stale}). Under
asynchronous scheduling, the host sizes the budget from
survivals produced two steps prior, one step older than the fresh ones the GPU
allocates it from
(\cref{sec:background}). This split is \dspark's
design~\citep[\S5.2]{cheng2026dspark}, and \sys keeps it. The n-gram drafter
suffers more from it than \dspark, because a match appears and vanishes
abruptly as the output enters and leaves a region it copies. The per-request
miscount of expected accepted drafts therefore has a standard deviation $\sigma$ of 2.2--3.4 for the
n-gram drafter and \sys, against 0.85--1.74 for \dspark. Requests miscount
independently, so the batch's shortfall grows only as $\sigma\sqrt{B}$ while its
expected accepted drafts grow as $B$, and the relative shortfall falls as
$1/\sqrt{B}$. On Qwen3-4B, stale sizing costs the n-gram drafter 7.2--8.9\% of
goodput at BS~4 and \sys 2.3--2.6\%, but at most 0.64\% and 0.30\% at BS~128
on NextCoder, where \av matters most. Every throughput number in this paper
already includes this loss. \Cref{app:stale} derives the $\sqrt{B}$ law, shows
that it predicts the measured shortfall, and explains why a shortfall costs
much less goodput than its size.

\subsection{Sampling at T{=}1 (Q7)}
\label{sec:eval-sampling}

\textbf{At T{=}1, the fastest lossless choice depends on the content.} At
T{=}1, each policy must be paired with a drafting mode under which it stays
lossless (\cref{sec:lossless}): greedy drafting for the proposed policy, either
mode for n-gram-first and the extension. On code editing, the proposed policy
with greedy drafting is the fastest in three of the four NextCoder cells
(\cref{tab:policy,tab:policy-full}; CIs exclude 1 in all four), 2.5--4.1\%
ahead of the better probabilistic alternative at BS~4 and 5.1--8.2\% at
BS~128. In the fourth cell, 40--45\% at BS~128, the extension with greedy
drafting is 1.6\% faster. Of the two alternatives, the extension, which appends the n-gram tail
to \dspark's unchanged chain, keeps most of the proposed policy's gain
(\cref{app:policy}). On chat the order reverses at BS~4: the proposed policy is 3.9\% slower
than the extension and 4.3\% slower than \dspark, both drafting
probabilistically, while at BS~128 all three are level.

\textbf{Probabilistic drafting is beneficial only where the next token is
uncertain.} A greedy draft token $\hat{x}$ is accepted with probability
$p(\hat{x})$ under the target's distribution $p$, while a token sampled from
\dspark's distribution $q$ is accepted with probability $\sum_x \min(p(x),
q(x))$, which is close to 1 whenever $q$ is close to $p$. On code editing,
probabilistic drafting does not raise \dspark's AL (6.1--6.7 in either
mode), so it only adds its cost: \dspark is 1.6--3.6\% slower with it. On
chat it raises \dspark's AL from 3.5 to 3.6 at BS~4, and \dspark is 2.9\%
faster with it. The drafting mode is a server setting, so a deployment at
T{=}1 should choose it according to its traffic.

\section{Related Work}
\label{sec:related}

\paragraph{Model-based drafters.}
Model-based drafters, surveyed by \citet{xia2024survey}, include separate small
models~\citep{leviathan2023fast}, heads on the target such as
Medusa~\citep{cai2024medusa}, token trees~\citep{miao2024specinfer},
multi-token prediction heads~\citep{gloeckle2024mtp} and
EAGLE~\citep{li2024eagle,li2024eagle2,li2025eagle3}. Parallel drafters such as
DFlash~\citep{chen2026dflash} and Domino~\citep{huang2026domino} emit a whole
block per forward pass. Any of them could fill the model-based slot given
calibrated confidences.

\paragraph{Model-free and retrieval drafters.}
Prompt Lookup Decoding~\citep{saxena2023pld} and LLMA~\citep{yang2023llma} copy
spans from the prompt; REST~\citep{he2024rest} retrieves from a datastore;
SuffixDecoding~\citep{oliaro2024suffix} and SAM Decoding~\citep{hu2024sam} index
prior outputs; Lookahead decoding~\citep{fu2024lookahead} collects n-grams from
the target's own parallel decoding. Blazedit~\citep{blazedit2025} speeds up a draft model with prompt
lookup for code editing, and Double~\citep{shen2026double} adds retrieval to
parallel speculative decoding. Any of these could fill \sys's model-free slot
given features and a fitted estimator (\cref{sec:estimator}).
Drafters that run on the host would also need a GPU or asynchronous
implementation (C3). Graft~\citep{shen2026graft}
fills pruned tree slots with retrieved candidates within each request, whereas
\sys chooses between whole chains across the batch at a shared price.

\paragraph{Speculative decoding for RL rollouts.}
SPEC-RL~\citep{liu2025specrl} verifies the previous epoch's rollout of the
same prompt under the current policy, with a lenience parameter that relaxes
acceptance and so changes the output distribution. RhymeRL~\citep{he2025rhymerl}
drafts from reward-aware suffix trees over each prompt's recent responses and
sets the draft length with an AIMD-inspired rule. DAS~\citep{shao2026das}
drafts from per-problem suffix trees over a sliding window of recent rollouts
and gives long trajectories larger draft budgets. Each uses one history drafter
with a heuristic draft length. \sys would instead combine such a drafter with a
model-based drafter under one calibrated, priced budget.

\paragraph{Adaptive verification.}
TurboSpec~\citep{liu2024turbospec} tunes the draft length per step for goodput,
and SpecDec++~\citep{huang2025specdecpp} and DISCO~\citep{mamou2024disco} set it
per request from predicted acceptance. PEARL~\citep{liu2025pearl} and
SpecBranch~\citep{shen2026specbranch} overlap drafting with verification, and
TALON~\citep{liu2026talon} grows a tree within a per-request token budget.
Across a batch, TETRIS~\citep{wu2025tetris} selects draft tokens under a fixed
verification capacity; \dspark~\citep{cheng2026dspark} and
D-cut~\citep{liu2026dcut} prune them under a profiled cost model; and
ECHO~\citep{hu2026echo} schedules tree depth and width under a global cap. All
decide how much of one drafter's chain to verify. \sys extends this
allocation to two drafters with different confidence sources.

\paragraph{Hybrid drafting.}
\citet{liu2026illusion} bound the gain of an oracle that picks between an
n-gram drafter and EAGLE per step, a choice that \sys makes with a deployable
policy.
SuffixDecoding~\citep{oliaro2024suffix} likewise pairs a model-free drafter
with a model-based one: per request, it takes the suffix-tree draft when its score, a
sum of continuation frequencies in the tree, exceeds a fixed threshold. This
rule is incompatible with adaptive verification in three ways. First, the score
counts how often a continuation appeared in past outputs, not how often the
target accepted it, so it cannot be ranked against \dspark's survivals in a
batch-wide budget. \sys instead fits acceptance probabilities
(\cref{sec:estimator}). Second, the fixed threshold does not rise with batch
size as verification becomes compute bound, while \sys prices slots by the
goodput-optimal budget
(\cref{sec:selection}). Third, its host-side suffix tree forces a CPU--GPU
synchronization every step, which breaks the zero-synchronization design of
vLLM's Model Runner V2 for asynchronous scheduling~\citep{vllm2026mrv2}. Its
vLLM implementation runs on the V1 model runner instead. \sys keeps every
per-step decision on the GPU (C3).

\section{Conclusion}
\label{sec:conclusion}

Model-based drafters perform well on general text, while n-gram drafters are
faster on near-copy edits. Yet most deployments
fix a single drafter, and adaptive verification can rank only drafts that carry
confidences. \sys runs both drafters on every step and gives n-gram drafts
calibrated acceptance estimates, fit once on general text. It prices each
verified slot at the goodput it would earn elsewhere in the batch: each request
keeps the chain whose slots are worth their price, and adaptive verification
allocates the budget over the chosen chains. Every per-step
decision runs on the GPU, which keeps asynchronous scheduling intact.

Instantiated with \dspark and an n-gram drafter in vLLM, \sys achieves
1.10--1.62\x the output throughput of \dspark on code editing at BS~4 and
1.04--1.27\x at BS~128, while staying at 0.97--1.01\x of it on chat at
$T{=}0$. A single deployment can therefore serve both editing and general
traffic, without partitioning GPUs into drafter-specific pools or forgoing the
editing speedup. Either slot can in principle hold other drafters. A natural
next step is RL post-training, where a history-based drafter built from earlier
rollouts could fill the model-free slot.

\bibliography{refs}
\bibliographystyle{plainnat}

\clearpage
\appendix
\label{appendixstart}
\onecolumn

\section{Notation}
\label{app:notation}

\begin{table}[H]
\centering
\small
\caption{Notation.}
\label{tab:notation}
\begin{tabular}{ll}
\toprule
Symbol & Meaning \\
\midrule
$B$ & batch size (number of running requests) \\
$K$, $L$ & \dspark chain length (7); maximum n-gram chain length (15) \\
$n$, $a$, $m$ & n-gram match length (5--15); runner-up agreement (0/1); copy length (capped at 256) \\
$c_k$, $h_k$ & conditional acceptance of position $k$: \dspark's confidence; \sys's n-gram estimate \\
$S_k$ & survival, $\prod_{j \le k}$ of the conditional acceptances \\
$b$, $b^*$ & verification budget (slots per step); budget chosen by \av \\
$C(b)$, $C_0$, $p$ & step cost; base cost; slot cost \\
$A(b)$, $G(b)$ & expected emitted tokens; goodput $A(b)/C(b)$ \\
$\lambda$ & slot price $G(b^*)\,p$ \\
$V$ & chain value $\sum_k \max(S_k - \lambda, 0)$ \\
\bottomrule
\end{tabular}
\end{table}

\section{Models and drafters}
\label{app:models}

\Cref{tab:models} lists the Hugging Face checkpoints of every target model and
its \dspark drafter.

\begin{table}[H]
\centering
\small
\caption{Target models and their \dspark drafters ($K{=}7$).
$^\dagger$Loaded with two \texttt{rope\_parameters} keys removed from its
configuration, which the installed Transformers version rejects. The weights
are unchanged.}
\label{tab:models}
\resizebox{\linewidth}{!}{%
\begin{tabular}{ll}
\toprule
Target model & \dspark drafter \\
\midrule
\texttt{Qwen/Qwen3-4B} & \texttt{deepseek-ai/dspark\_qwen3\_4b\_block7} \\
\texttt{Qwen/Qwen3-8B} & \texttt{deepseek-ai/dspark\_qwen3\_8b\_block7} \\
\texttt{google/gemma-4-12B-it} & \texttt{deepseek-ai/dspark\_gemma4\_12b\_block7}$^\dagger$ \\
\texttt{CohereLabs/North-Mini-Code-1.0-w4a16} & \texttt{sdougbrown/North-Mini-Code-1.0-dspark} \\
\texttt{meta-llama/Llama-3.1-8B-Instruct} & \begin{tabular}[t]{@{}l@{}}\texttt{weifanjiang/llama3.1-8b-it.speculators.}\\\texttt{\ \   dspark-ce01tv09-bs8-conf-swa}\end{tabular} \\
\bottomrule
\end{tabular}}
\end{table}

\section{Workloads}
\label{app:workloads}

\Cref{tab:workloads} lists the size of each workload, the prompts per run and
the output limit.

\begin{table}[H]
\centering
\small
\caption{Workloads. Prompts per run at BS~4/32/128; small sets are
oversampled to that count. Bucketed sets run each bucket with at least that
many prompts.}
\label{tab:workloads}
\setlength{\tabcolsep}{3pt}
\begin{tabular}{lrrr}
\toprule
Workload & Size & Prompts per run & Max out \\
\midrule
Blazedit 5k / 10k & 21 / 19 & 64 / 256 / 512 & 4{,}096 \\
CanItEdit         & 30      & 64 / 256 / 512 & per prompt \\
NextCoder         & 4{,}000 & 14 buckets, 0--70\% & 4{,}096 \\
EditBench         & 520     & 5 buckets, 0--25\% & 4{,}096 \\
InstructCoder     & 3{,}200 & 200 / 800 / 3{,}200 & 200 \\
MT-Bench          & 80      & 160 / 640 / 1{,}280 & 256 \\
\bottomrule
\end{tabular}
\end{table}

\section{Estimator coefficients}
\label{app:coef}

\begin{table}[H]
\centering
\small
\caption{Fitted estimator coefficients (\cref{eq:estimator}), fit on the
SPEED-Bench fitting half. Position intercepts $\beta_1, \dots, \beta_{15}$.}
\label{tab:coef}
\setlength{\tabcolsep}{4pt}
\resizebox{\linewidth}{!}{%
\begin{tabular}{lrrrrrrrrrrrrrrr}
\toprule
 & $\beta_1$ & $\beta_2$ & $\beta_3$ & $\beta_4$ & $\beta_5$ & $\beta_6$ & $\beta_7$ & $\beta_8$ & $\beta_9$ & $\beta_{10}$ & $\beta_{11}$ & $\beta_{12}$ & $\beta_{13}$ & $\beta_{14}$ & $\beta_{15}$ \\
\midrule
Qwen3-4B & $-$0.220 & 0.155 & 0.506 & 0.683 & 0.788 & 0.697 & 0.766 & 0.721 & 0.631 & 0.658 & 0.761 & 0.803 & 0.762 & 0.862 & 0.811 \\
Qwen3-8B & $-$0.414 & $-$0.006 & 0.246 & 0.457 & 0.551 & 0.372 & 0.613 & 0.496 & 0.501 & 0.448 & 0.516 & 0.573 & 0.554 & 0.479 & 0.744 \\
Gemma~4 12B & $-$0.630 & $-$0.227 & $-$0.087 & 0.086 & 0.456 & 0.263 & 0.340 & 0.541 & 0.401 & 0.443 & 0.566 & 0.678 & 0.759 & 0.779 & 0.932 \\
North-Mini-Code & $-$0.426 & $-$0.204 & $-$0.045 & 0.044 & 0.175 & 0.084 & $-$0.051 & $-$0.137 & $-$0.417 & $-$0.586 & $-$0.582 & $-$0.581 & $-$0.478 & $-$0.308 & $-$0.154 \\
Llama-3.1-8B & $-$0.070 & 0.287 & 0.544 & 0.726 & 0.950 & 0.958 & 0.897 & $-$0.420 & 0.607 & 0.485 & 0.602 & 0.537 & 0.611 & 0.362 & 0.762 \\
\bottomrule
\end{tabular}}

\vspace{0.6em}
\begin{tabular}{lrrrrrrr}
\toprule
Model & $w_{\text{first}}$ & $w_{\text{rest}}$ & $\delta$ & $\ell_{\text{first}}$ & $\ell_{\text{rest}}$ & $\kappa_{\text{first}}$ & $\kappa_{\text{rest}}$ \\
\midrule
Qwen3-4B & $-$0.0893 & $-$0.0001 & 1.2455 & 1.1998 & 0.1322 & 2.0118 & 1.8333 \\
Qwen3-8B & $-$0.1301 & 0.0303 & 1.3938 & 1.2694 & $-$0.0385 & 2.5317 & 1.7287 \\
Gemma~4 12B & $-$0.0173 & 0.0683 & 1.1741 & 1.1134 & 0.1844 & 2.1470 & 1.5580 \\
North-Mini-Code & $-$0.1422 & $-$0.0757 & 2.0234 & 1.3238 & 0.9809 & 1.6695 & 1.6321 \\
Llama-3.1-8B & $-$0.1248 & 0.0233 & 1.4169 & 1.9266 & 0.0714 & 1.3100 & 1.5100 \\
\bottomrule
\end{tabular}
\end{table}

The agreement weight is large and positive ($\delta = 1.25$ for Qwen3-4B), and
a 256-token copy adds $\kappa_{\text{first}} \ln(256/15) \approx 5.7$ logits to
the first position of a full match.

\paragraph{Fitting data (Qwen3-4B / Qwen3-8B).} 342{,}593 / 328{,}854 output
tokens; 749 / 709 requests with at least one match; 19{,}699 / 17{,}131 matched
steps, of which 4{,}904 / 4{,}022 are full matches; 135{,}518 / 110{,}853
graded positions, of which 120{,}958 / 97{,}981 were accepted. The median copy
length is 85 / 106 tokens, and 20\% / 26\% of copies reach the 256-token cap.

\section{Building the estimator up one term at a time}
\label{app:ladder}

Giving the n-gram drafter any calibrated survival is what lets it use \av,
and that alone captures most of the gain. With one shared intercept fitted on
SPEED-Bench and no match features, the n-gram drafter with \av reaches
0.973\x [0.966, 0.980] of the full estimator's output throughput on pooled
NextCoder at BS~128 (\cref{tab:ladder}). Verifying every draft is 1.89\x
slower than the full estimator on the same buckets (separate runs, 3 to 5
repeats), so one shared intercept keeps about 94\% of the full estimator's
gain over verifying every draft.

\textbf{Setup.} We serve the n-gram drafter with \av, where every survival
\av uses comes from the estimator, with six estimators of increasing
richness: (1) one intercept $\beta$ shared by all positions; (2) per-position
intercepts $\beta_k$; (3) $\beta_k$ and match length $w$; (4) $\beta_k$ and
copy length $\kappa$; (5) $\beta_k$, $\kappa$ and agreement $\delta$; and
(6) the full estimator of \cref{eq:estimator}. Estimators 3 and 4 each add
one feature to estimator 2. Each estimator is refit on the same SPEED-Bench
fitting half (\cref{sec:fitting}), with the omitted terms fixed at zero; in
estimators 3--5 each feature has one coefficient shared by the first and
later tokens. \Cref{tab:ladder-coef} lists the coefficients of estimators
1--5, and \cref{tab:coef} those of estimator 6. Even the single shared
intercept is a fitted estimator: $\beta = 2.117$ predicts an acceptance of
0.893 at every position, so survival still decays with position and \av
still trims long chains. Qwen3-4B, NextCoder 15--20\% and 40--45\%, BS~128,
8 repeats, each on a fresh server, with the six estimators interleaved
within each repeat.

\textbf{Copy length adds the most, and the other terms add little.} Measured
from per-position intercepts (0.978\x), copy length adds 1.8\% (0.996\x
[0.987, 1.003]), and every estimator with it is level with the full one.
Match length adds 0.6\% (0.984\x), and per-position intercepts add 0.5\% over
one shared intercept. These two steps are within noise of each other, and
their order differs between buckets: per-position intercepts help on
15--20\% (0.973\x to 0.988\x) but not on 40--45\% (0.973\x to 0.971\x), and
match length does the reverse (0.976\x and 0.990\x). Agreement, the
full-match bonus and the first-token split add nothing measurable on top of
copy length (0.996\x [0.991, 1.001] with agreement added). The match features
still matter for calibration: copy length lowers the estimator's calibration
error on EditBench from 0.132 to 0.032, and agreement separates first-token
acceptance of 24\% from 81\% (\cref{sec:eval-components}).

\textbf{The estimator chooses which slots to verify; it does not raise
acceptance.} AL stays within 3.65--3.68 on 15--20\% and 2.29--2.33 on
40--45\% across all six estimators, so the throughput differences come from
which slots \av verifies, not from how many drafts are accepted.

\textbf{Scope.} On these two buckets, intercepts alone cost 2.2\%, less than
the 4.1\% pooled over all NextCoder buckets in \cref{sec:eval-components}.
Inside \sys the same estimators stay level with the full one (intercepts
alone: 0.985\x [0.959, 1.007], 5 repeats).

\textbf{Coefficients.} Estimator 2's intercepts rise from 1.158 at the first
position to 3.046 at the fifteenth, an acceptance of 0.76 rising to 0.95,
which is the profile of Observation~2 (\cref{tab:ladder-coef}). Adding a
feature lowers the intercepts, because the feature then explains part of
the acceptance. Fitted alone, match length has a positive weight
($w = 0.196$): longer matches are accepted more often. In the full
estimator, its first-token weight is negative ($w_{\text{first}} = -0.089$,
\cref{tab:coef}), because the full-match bonus $\ell$ and copy length
$\kappa$ take over the effect of long matches, so the two signs do not
conflict.

\begin{table}[H]
  \centering
  \small
  \caption{A single shared intercept already reaches 97\% of the full
  estimator's throughput, and copy length closes most of the rest. Output
  throughput of the n-gram drafter with \av relative to the full estimator
  (higher is better), Qwen3-4B, NextCoder, BS~128, 8 repeats, 95\% CI for
  the pooled buckets. Estimators 3 and 4 each add one feature to estimator
  2. Symbols
  as in \cref{eq:estimator}: $\beta$ one shared intercept, $\beta_k$
  per-position intercepts, $w$ match length, $\kappa$ copy length, $\delta$
  agreement.}
  \label{tab:ladder}
  \begin{tabular}{@{}clrrr@{}}
    \toprule
    & Terms & 15--20\% & 40--45\% & Pooled [95\% CI] \\
    \midrule
    1 & $\beta$ & 0.973 & 0.973 & 0.973 [0.966, 0.980] \\
    2 & $\beta_k$ & 0.988 & 0.971 & 0.978 [0.971, 0.985] \\
    3 & $\beta_k$, $w$ & 0.976 & 0.990 & 0.984 [0.978, 0.990] \\
    4 & $\beta_k$, $\kappa$ & 0.999 & 0.993 & 0.996 [0.987, 1.003] \\
    5 & $\beta_k$, $\kappa$, $\delta$ & 0.995 & 0.998 & 0.996 [0.991, 1.001] \\
    6 & all (\cref{eq:estimator}) & 1 & 1 & 1 \\
    \bottomrule
  \end{tabular}
\end{table}

\begin{table}[H]
\centering
\small
\caption{Coefficients of estimators 1--5 in \cref{tab:ladder} (Qwen3-4B,
  SPEED-Bench fitting half); estimator 6 is the full estimator in
  \cref{tab:coef}. Estimators 3--5 use one coefficient for the first and
  later tokens. Copy length enters only for full matches, as in
  \cref{eq:estimator}. A dash marks a term the estimator omits.}
\label{tab:ladder-coef}
\setlength{\tabcolsep}{4pt}
\resizebox{\linewidth}{!}{%
\begin{tabular}{crrrrrrrrrrrrrrrrrr}
\toprule
 & $\beta_1$ & $\beta_2$ & $\beta_3$ & $\beta_4$ & $\beta_5$ & $\beta_6$ & $\beta_7$ & $\beta_8$ & $\beta_9$ & $\beta_{10}$ & $\beta_{11}$ & $\beta_{12}$ & $\beta_{13}$ & $\beta_{14}$ & $\beta_{15}$ & $w$ & $\kappa$ & $\delta$ \\
\midrule
1 & \multicolumn{15}{c}{2.117 at every position} & -- & -- & -- \\
2 & 1.158 & 1.702 & 2.104 & 2.325 & 2.466 & 2.420 & 2.536 & 2.544 & 2.515 & 2.605 & 2.757 & 2.854 & 2.877 & 3.036 & 3.046 & -- & -- & -- \\
3 & 0.793 & 1.293 & 1.658 & 1.845 & 1.962 & 1.882 & 1.960 & 1.923 & 1.841 & 1.876 & 1.981 & 2.025 & 1.992 & 2.098 & 2.053 & 0.196 & -- & -- \\
4 & 0.860 & 1.345 & 1.700 & 1.880 & 1.987 & 1.897 & 1.967 & 1.922 & 1.834 & 1.859 & 1.956 & 1.994 & 1.950 & 2.045 & 1.988 & -- & 2.070 & -- \\
5 & $-$0.247 & 0.144 & 0.495 & 0.671 & 0.776 & 0.685 & 0.754 & 0.709 & 0.618 & 0.646 & 0.749 & 0.791 & 0.750 & 0.850 & 0.799 & -- & 2.014 & 1.261 \\
\bottomrule
\end{tabular}}
\end{table}

\section{Cost model}
\label{app:cost}

For reproducibility we describe how \av's step-cost model
(\cref{sec:selection,sec:budget}) is built. It is profiled once at startup. The
step cost is $C(b) = C_0 + p\,b$, and only the ratio of $p$ to $C_0$ affects
decisions.
\begin{itemize}
  \setlength\itemsep{0pt}
  \item The \emph{slot cost} $p$ for $R$ requests is the time difference between
  a verification pass with every query slot filled and a pass with one token per
  request, divided by the number of slots added. It is tabulated per request
  count.
  \item The \emph{base cost} $C_0 = T(N) + C_{\mathrm{KV}}$ is a token-count
  term $T(N)$ for the $N$ non-draft tokens plus a KV-read term $C_{\mathrm{KV}}$
  proportional to the total context. Both come from two profiling passes, with
  and without synthetic context.
  \item Above the CUDA graph capture limit, a step runs eagerly. The jump from
  graphed to eager execution is not linear in $b$, so for $N + b$ above the
  limit the cost is
  \begin{equation}
  C(b) = \max\bigl(C_0 + p\,b,\; T(N + b) + C_{\mathrm{KV}}\bigr),
  \label{eq:eager}
  \end{equation}
  where $T(N + b)$ is the profiled cost of an eager step with $N + b$ tokens.
\end{itemize}
All of this is host-side arithmetic over arrays the scheduler already holds.

\section{Chain value: a worked example}
\label{app:value}

The slot price $\lambda = G(b^*)\,p$ (\cref{eq:price}) depends on the batch
size and the hardware through the step costs. For Qwen3-8B on NextCoder on an
H100 GPU (\cref{sec:eval-setup}), it is about 0.04 at BS~4 and about 0.3 at
BS~128.

\Cref{tab:value-example} shows a case where ranking chains by expected accepted
tokens, $\sum_k S_k$, keeps the n-gram chain and the value $V$
(\cref{eq:value}) keeps \dspark's. Survivals are averaged over steps for
Qwen3-8B on NextCoder 45--50\% at BS~128, where $\lambda = 0.309$. The n-gram
chain's lead in $\sum_k S_k$ comes entirely from its 6 slots below $\lambda$,
which add 1.52 expected tokens, about 0.25 per slot, while each slot would earn
0.309 elsewhere in the batch. On the 15--20\% bucket ($\lambda = 0.288$), the
n-gram chain is stronger and also has the higher value, 4.36 against 4.04. The
value rule thus drops the n-gram chain only where its slots are worth less than
their price, not at every large batch size.

\begin{table}[H]
\centering
\small
\caption{Chain value at BS~128 (Qwen3-8B, NextCoder 45--50\%,
$\lambda = 0.309$). $\sum_k S_k$ prefers the n-gram chain, whereas $V$
prefers \dspark's.}
\label{tab:value-example}
\begin{tabular}{lrrrr}
\toprule
Chain & $\sum_k S_k$ & slots $S_k > \lambda$ & their $\sum S_k$ & $V$ \\
\midrule
n-gram & 5.87 & 9 & 4.35 & 1.57 \\
\dspark & 5.58 & 7 & 5.58 & 3.42 \\
\bottomrule
\end{tabular}
\end{table}

\section{Losslessness}
\label{app:lossless}

This appendix expands \cref{sec:lossless}.

\textbf{Greedy drafting is lossless at any temperature.} At target temperature
0, any draft only changes speed. At higher temperatures, with greedy drafting,
both chains and every selection and allocation decision are functions of the
context alone. The target verifies each draft token as a point mass with
standard rejection sampling, so the output distribution is exactly the target's.
D-cut makes the same argument for pruning deterministic
drafts~\citep[App.~A]{liu2026dcut}.

\textbf{Probabilistic drafting needs a restricted policy.} When \dspark samples
its drafts from its own distribution $q$, verifying slot $i$ against $q_i$
preserves the target's distribution only if the decision to verify slot $i$, and
how, is independent of the token drawn at slot $i$ given the tokens before
it~\citep{cheng2026dspark}. Point-mass drafts fixed before verification,
independently of the sampled tokens, are exempt. D-cut gives a one-token example
of the bias when this fails~\citep[App.~B]{liu2026dcut}. Our proposed policy
(\cref{sec:selection}) violates the condition in two ways. First, the contains
rule compares \dspark's sampled tokens with the n-gram chain, so whether
\dspark's chain is kept depends on the tokens \dspark drew. Second, the value
rule scores \dspark's chain from its confidences $c_k$, and \dspark's confidence
head computes $c_k$ from the token it sampled at position $k-1$. A sampled token
that makes \dspark confident about the next positions raises its chain's value
$V$ and keeps the chain, while one that makes it unsure lowers $V$ and hands the slots
to the n-gram chain. Whether \dspark's token at slot $i$ is verified then
depends on that token itself, which skews the output toward tokens after which
\dspark is confident.

\textbf{Two lossless alternatives.} For probabilistic drafting we therefore
consider two simpler policies. (1)~\emph{N-gram-first} takes the n-gram chain
whenever it matched and \dspark's chain otherwise. A model-free chain is fixed
before verification and does not depend on \dspark's sampled tokens
(\cref{sec:overview}), so the choice never reads \dspark's samples. (2)~The
\emph{extension} policy always verifies \dspark's chain against $q$ and appends
the n-gram tail only when \dspark's whole chain is a prefix of the n-gram chain.
Slots 1--7 are unchanged. A tail slot is reached only if all 7 \dspark tokens
are accepted, so the extension decision reads only tokens before it, and a
per-slot flag tells the rejection sampler to verify tail slots as point masses.
Tail survivals continue from \dspark's last survival, so a tail slot never
outranks \dspark's own slots. \Cref{tab:lossless} summarizes where each policy
is lossless. Since greedy drafting is lossless at any temperature, a deployment
that samples at T{=}1 can either keep the proposed policy with greedy drafting
or switch to an alternative with probabilistic drafting. Which is faster
depends on the workload (\cref{sec:eval-sampling}).

\begin{table}[H]
\centering
\small
\caption{Where each chain-selection policy preserves the target's output
distribution exactly.}
\label{tab:lossless}
\setlength{\tabcolsep}{4pt}
\begin{tabular}{llccc}
\toprule
Target & Drafting & Proposed & N-gram-first & Extension \\
\midrule
$T = 0$ & any           & yes & yes & yes \\
$T > 0$ & greedy        & yes & yes & yes \\
$T > 0$ & probabilistic & no  & yes & yes \\
\bottomrule
\end{tabular}
\end{table}

\section{Where \dspark's drafter is weak}
\label{sec:eval-ood}

In these cases \dspark's drafter accepts few tokens per step, so \dspark itself
is slow and the speedups of \sys over it are large partly for that reason. A
better-trained drafter would shrink them, so we keep them out of our headline
claims.

\textbf{North-Mini-Code.} On Blazedit and CanItEdit at BS~4, \dspark
accepts 2.3--3.2 tokens per step, against 5.7--6.9 for \sys
(\cref{fig:al}). \sys achieves 1.56--2.50\x over \dspark on Blazedit and
CanItEdit at BS~4 and 1.27--1.62\x at BS~128 (\cref{fig:main}), 1.28\x
and 1.23\x on pooled NextCoder, and 1.16\x and 1.15\x on InstructCoder at
BS~4 and 128.

\textbf{Llama-3.1-8B and long contexts.} \Cref{tab:ood} reports Llama-3.1-8B with
a community-trained \dspark drafter (AL 2.0--3.2 on editing) and EditBench
requests with more than 6{,}000 tokens of context, where the in-distribution
drafters' AL falls (from about 6.5 below 6k tokens to 2.7 above 12k for
Qwen3-4B and Qwen3-8B). \sys achieves up to 7.58\x over \dspark at BS~4 and 3.87\x at BS~128,
and Llama on MT-Bench is level (0.98--1.01\x).

\begin{table}[H]
\centering
\small
\caption{Out of distribution for the \dspark drafter: large gains that reflect
weak drafters. Left: output-throughput speedup of \sys over \dspark. Right:
acceptance length (tokens per verification step) of \dspark / \sys; NextCoder's
is the mean over its edit-distance buckets. EB-LC: EditBench requests with more
than 6{,}000 tokens of context.}
\label{tab:ood}
\setlength{\tabcolsep}{4pt}
\begin{tabular}{llrrrccc}
\toprule
 & & \multicolumn{3}{c}{Speedup} & \multicolumn{3}{c}{AL, \dspark / \sys} \\
\cmidrule(lr){3-5} \cmidrule(lr){6-8}
Model & Workload & BS 4 & BS 32 & BS 128 & BS 4 & BS 32 & BS 128 \\
\midrule
\multirow{5}{*}{Llama-3.1-8B}
  & Blazedit 5k       & 1.97 & 1.71 & 1.41 & 2.4 / 5.1 & 2.4 / 5.0 & 2.4 / 4.6 \\
  & Blazedit 10k      & 2.96 & 2.32 & 1.96 & 2.1 / 6.3 & 2.0 / 6.3 & 2.0 / 6.0 \\
  & CanItEdit         & 1.84 & 1.49 & 1.30 & 3.2 / 6.2 & 3.2 / 5.9 & 3.0 / 5.2 \\
  & NextCoder         & 1.69 & 1.60 & 1.46 & 2.9 / 5.5 & 2.9 / 5.4 & 2.8 / 4.9 \\
  & InstructCoder     & 1.22 & 1.12 & 1.07 & 3.3 / 4.1 & 3.1 / 3.9 & 2.8 / 3.6 \\
\midrule
\multirow{4}{*}{EB-LC}
  & Qwen3-4B          & 3.99 & 3.26 & 2.76 & 3.3 / 14.4 & 3.3 / 14.3 & 3.2 / 14.2 \\
  & Qwen3-8B          & 4.07 & 2.77 & 2.49 & 3.2 / 14.6 & 3.2 / 14.6 & 3.2 / 14.4 \\
  & Gemma 4 12B       & 2.24 & 1.73 & 1.51 & 5.8 / 14.2 & 5.6 / 14.0 & 5.4 / 13.9 \\
  & Llama-3.1-8B      & 7.58 & 5.09 & 3.87 & 1.6 / 14.3 & 1.6 / 14.3 & 1.6 / 14.1 \\
\bottomrule
\end{tabular}
\end{table}

\section{Values of Figure~\ref{fig:teaser}}
\label{app:teaser}

\begin{table}[H]
\centering
\small
\caption{Output-throughput speedup over vanilla decoding on NextCoder by
edit-distance bucket (Qwen3-4B, mean of 3 repeats; higher is better), the
values plotted in \cref{fig:teaser}. n-gram: every draft verified, plotted at
BS~128 only. The last row pools all buckets and is not plotted.}
\label{tab:teaser}
\begin{tabular}{lrrrrrrr}
\toprule
 & \multicolumn{3}{c}{BS~4} & \multicolumn{4}{c}{BS~128} \\
\cmidrule(lr){2-4}\cmidrule(lr){5-8}
Edit distance & \dspark & n-gram + \av & \sys & \dspark & n-gram & n-gram + \av & \sys \\
\midrule
0--5\% & 5.00 & 7.42 & 7.98 & 2.88 & 3.14 & 3.68 & 3.57 \\
5--10\% & 4.87 & 5.07 & 7.11 & 2.69 & 2.20 & 3.00 & 3.30 \\
10--15\% & 4.78 & 4.06 & 6.52 & 2.48 & 1.60 & 2.54 & 2.84 \\
15--20\% & 4.61 & 3.31 & 5.96 & 2.49 & 1.47 & 2.51 & 2.91 \\
20--25\% & 4.54 & 2.88 & 5.65 & 2.35 & 1.12 & 1.99 & 2.54 \\
25--30\% & 4.53 & 2.64 & 5.43 & 2.41 & 1.11 & 2.06 & 2.65 \\
30--35\% & 4.42 & 2.34 & 5.09 & 2.39 & 1.00 & 1.96 & 2.57 \\
35--40\% & 4.40 & 2.22 & 5.03 & 2.45 & 0.95 & 1.88 & 2.58 \\
40--45\% & 4.29 & 2.10 & 4.83 & 2.29 & 0.84 & 1.73 & 2.42 \\
45--50\% & 4.34 & 2.06 & 4.87 & 2.50 & 0.89 & 1.80 & 2.63 \\
50--55\% & 4.32 & 1.97 & 4.81 & 2.27 & 0.81 & 1.72 & 2.50 \\
55--60\% & 4.17 & 1.88 & 4.53 & 2.37 & 0.78 & 1.63 & 2.50 \\
60--65\% & 4.30 & 1.89 & 4.66 & 2.48 & 0.79 & 1.69 & 2.53 \\
65--70\% & 4.47 & 1.80 & 4.75 & 2.68 & 0.83 & 1.74 & 2.86 \\
\midrule
Pooled & 4.53 & 2.70 & 5.52 & 2.49 & 1.07 & 2.04 & 2.75 \\
\bottomrule
\end{tabular}
\end{table}

\section{Speedup per workload}
\label{app:main-all}

\Cref{tab:main-all} lists the output-throughput speedup of \sys over \dspark
for every model, workload and batch size in \cref{fig:main}, including the
pooled NextCoder and EditBench cells.

\begin{table}[H]
\centering
\small
\caption{Output-throughput speedup of \sys over \dspark (mean of 3 repeats;
$^\circ$: 95\% CI includes 1, level). NextCoder and EditBench are pooled over
their edit-distance buckets; EditBench excludes requests with more than
6{,}000 tokens of context. North-Mini-Code's \dspark drafter is weak
(\cref{sec:eval-ood}), so it is not part of the headline ranges.}
\label{tab:main-all}
\resizebox{\linewidth}{!}{%
\begin{tabular}{llrrrrrrr}
\toprule
Model & BS & Blazedit 5k & Blazedit 10k & CanItEdit & NextCoder & EditBench & InstructCoder & MT-Bench \\
\midrule
\multirow{3}{*}{Qwen3-4B} & 4 & 1.34 & 1.50 & 1.35 & 1.24 & 1.62 & 1.14 & 0.99$^\circ$ \\
 & 32 & 1.16 & 1.31 & 1.16 & 1.11 & 1.37 & 1.06 & 0.99 \\
 & 128 & 1.13 & 1.26 & 1.12 & 1.10 & 1.27 & 0.99 & 0.98 \\
\midrule
\multirow{3}{*}{Qwen3-8B} & 4 & 1.31 & 1.45 & 1.33 & 1.19 & 1.59 & 1.11 & 1.00$^\circ$ \\
 & 32 & 1.13 & 1.28 & 1.11 & 1.08 & 1.26 & 1.02 & 1.00$^\circ$ \\
 & 128 & 1.08 & 1.24 & 1.08 & 1.05 & 1.19 & 0.98 & 0.99 \\
\midrule
\multirow{3}{*}{Gemma~4 12B} & 4 & 1.30 & 1.40 & 1.26 & 1.10 & 1.52 & 1.08 & 1.01 \\
 & 32 & 1.16 & 1.25 & 1.02$^\circ$ & 1.03 & 1.15 & 1.03 & 1.01 \\
 & 128 & 1.08 & 1.13 & 1.05 & 1.04 & 1.12 & 0.98 & 1.00$^\circ$ \\
\midrule
\multirow{3}{*}{North-Mini-Code} & 4 & 1.92 & 2.50 & 1.56 & 1.28 & 2.69 & 1.16 & 0.99 \\
 & 32 & 1.68 & 2.06 & 1.34 & 1.28 & 2.05 & 1.10 & 1.00$^\circ$ \\
 & 128 & 1.44 & 1.62 & 1.27 & 1.23 & 1.69 & 1.15 & 1.00$^\circ$ \\
\bottomrule
\end{tabular}}
\end{table}

\section{Acceptance length per workload}
\label{app:al}

\begin{figure}[H]
  \centering
  \includegraphics[width=0.85\textwidth]{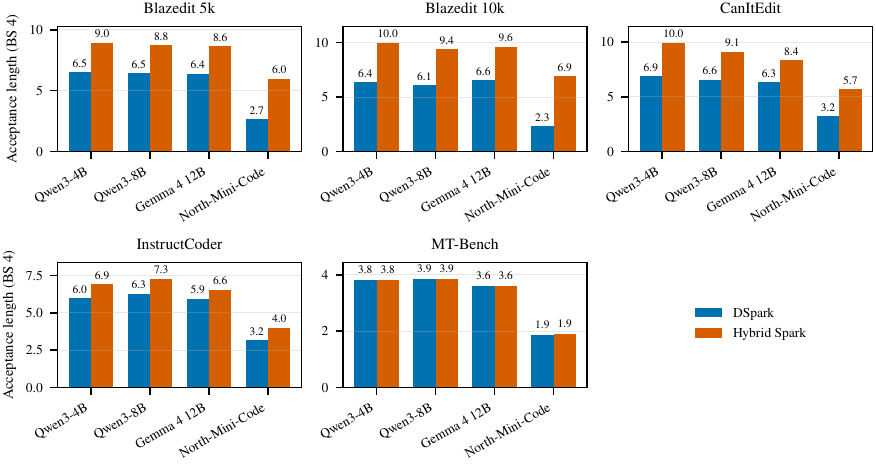}
  \caption{\sys accepts more tokens per step than \dspark on every editing
  workload and the same on MT-Bench. Acceptance length (tokens per
  verification step, higher is better) at BS~4, mean of 3 repeats.}
  \label{fig:al}
\end{figure}

\Cref{fig:al} shows the AL of \dspark and \sys at BS~4 on the five
single-file workloads of \cref{fig:main}.

\section{EditBench by edit distance}
\label{app:edit-eb}

\begin{figure}[H]
  \centering
  \includegraphics[width=0.5\textwidth]{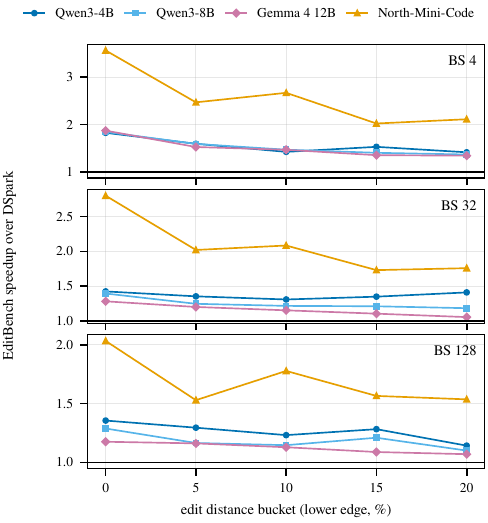}
  \caption{Output-throughput speedup of \sys over \dspark per EditBench bucket,
  without requests over 6{,}000 tokens of context (higher is better). Mean of 3
  repeats.}
  \label{fig:edit-eb}
\end{figure}

\Cref{fig:edit-eb} breaks EditBench down by edit distance, as the last three
panels of \cref{fig:main} do for NextCoder. As in the main results, requests
with more than 6{,}000 tokens of context are excluded from every bucket. The
available \dspark drafters were likely trained on little data this long, and
their AL falls from about 6.5 below 6k tokens to 2.7 above 12k
(\cref{sec:eval-ood}). \dspark is then slow and the gains of \sys over it are
inflated, so we leave these requests out. As on
NextCoder, the gain falls with edit distance for the three in-distribution
models: at BS~4 from 1.82--1.87\x at
0--5\% to 1.34--1.41\x at 20--25\%, and at BS~128 from 1.18--1.36\x to
1.07--1.14\x.

\section{All policy cells}
\label{app:policy}

\textbf{Of the alternatives, the extension keeps most of the proposed policy's
gain.} At T{=}0 it is within 5\% of the proposed policy's throughput, and on
40--45\% it is within 0.2\% of it or above it in every regime. N-gram-first is
0.8\% and 2.3\% faster than the extension on 15--20\% at T{=}1 but slower than
\dspark on 40--45\% at BS~128. \Cref{tab:policy-full,tab:policy-mtb} list all
four regimes for NextCoder and MT-Bench, each over \dspark with greedy
drafting, as in \cref{tab:policy}.

\begin{table}[H]
\centering
\small
\caption{Output-throughput speedup over \dspark with greedy drafting at the
same temperature, all four sampling regimes (Qwen3-4B, NextCoder, 3 repeats),
as in \cref{tab:policy}. T0p: T{=}0 with probabilistic drafting; T1g: T{=}1
with greedy drafting; \dspark (prob.): \dspark with probabilistic drafting.
$^\circ$CI includes 1. --: not lossless, not run.}
\label{tab:policy-full}
\begin{tabular}{llrrrrrrrr}
\toprule
 & & \multicolumn{4}{c}{NextCoder 15--20\%} & \multicolumn{4}{c}{NextCoder 40--45\%} \\
\cmidrule(lr){3-6}\cmidrule(lr){7-10}
BS & Policy & T0g & T0p & T1g & T1p & T0g & T0p & T1g & T1p \\
\midrule
\multirow{4}{*}{4}
  & Proposed       & 1.301 & 1.304 & 1.282 & -- & 1.127 & 1.123 & 1.126 & -- \\
  & N-gram-first   & 1.269 & 1.274 & 1.244 & 1.231 & 1.067 & 1.067 & 1.070 & 1.056 \\
  & Extension      & 1.269 & 1.271 & 1.241 & 1.222 & 1.128 & 1.131 & 1.124 & 1.098 \\
  & \dspark (prob.) &  & 0.994$^\circ$ &  & 0.976 &  & 0.994$^\circ$ &  & 0.974 \\
\midrule
\multirow{4}{*}{128}
  & Proposed       & 1.163 & 1.159 & 1.178 & -- & 1.053 & 1.051 & 1.067 & -- \\
  & N-gram-first   & 1.104 & 1.106 & 1.119 & 1.089 & 0.976 & 0.976 & 0.970$^\circ$ & 0.962$^\circ$ \\
  & Extension      & 1.109 & 1.111 & 1.077$^\circ$ & 1.065 & 1.061 & 1.062 & 1.084 & 1.015$^\circ$ \\
  & \dspark (prob.) &  & 0.996$^\circ$ &  & 0.964$^\circ$ &  & 0.993 &  & 0.984$^\circ$ \\
\bottomrule
\end{tabular}
\end{table}

\begin{table}[H]
\centering
\small
\caption{As \cref{tab:policy-full}, on MT-Bench (Qwen3-4B, 3 repeats).}
\label{tab:policy-mtb}
\begin{tabular}{llrrrr}
\toprule
BS & Policy & T0g & T0p & T1g & T1p \\
\midrule
\multirow{4}{*}{4}
  & Proposed       & 0.989 & 0.981 & 0.985 & -- \\
  & N-gram-first   & 0.980 & 0.976 & 0.982 & 1.015 \\
  & Extension      & 0.997$^\circ$ & 0.990$^\circ$ & 1.000$^\circ$ & 1.025 \\
  & \dspark (prob.) &  & 1.000$^\circ$ &  & 1.029 \\
\midrule
\multirow{4}{*}{128}
  & Proposed       & 0.968 & 0.964 & 0.971 & -- \\
  & N-gram-first   & 0.970 & 0.964 & 0.980$^\circ$ & 0.965 \\
  & Extension      & 0.953 & 0.974 & 0.979$^\circ$ & 0.977$^\circ$ \\
  & \dspark (prob.) &  & 0.983 &  & 0.987$^\circ$ \\
\bottomrule
\end{tabular}
\end{table}

\textbf{SPEED-Bench at T{=}1.} \Cref{tab:policy-sb} repeats the T{=}1 regimes
on the 1k and 2k SPEED-Bench prompts of \cref{app:sbctx}, with the same
serving setup. The largest speedups are on low-entropy prompts at BS~4:
1.06\x for the proposed policy, 1.07\x for n-gram-first and 1.04\x for the
extension. With the better drafting mode per policy, the worst case is 0.97\x
for the proposed policy, 0.98\x for the extension and 0.96\x for
n-gram-first, all on mixed or high-entropy prompts. Outside low-entropy
prompts at BS~4, the fastest lossless choice is 1.00--1.02\x, and in 6 of
these 10 cells it is \dspark itself. \Cref{fig:sbctx} gives the proposed
policy at T{=}0 on the same prompts.

\begin{table}[H]
\centering
\small
\setlength{\tabcolsep}{4.5pt}
\caption{As \cref{tab:policy-full}, at T{=}1 on held-out SPEED-Bench prompts
with 1k and 2k tokens of context, per category (low entropy, mixed, high
entropy; Qwen3-4B, FP8 KV cache, 3 repeats).}
\label{tab:policy-sb}
\resizebox{\linewidth}{!}{%
\begin{tabular}{llrrrrrrrrrrrr}
\toprule
 & & \multicolumn{6}{c}{1k context} & \multicolumn{6}{c}{2k context} \\
\cmidrule(lr){3-8}\cmidrule(lr){9-14}
 & & \multicolumn{2}{c}{Low} & \multicolumn{2}{c}{Mixed} & \multicolumn{2}{c}{High}
 & \multicolumn{2}{c}{Low} & \multicolumn{2}{c}{Mixed} & \multicolumn{2}{c}{High} \\
\cmidrule(lr){3-4}\cmidrule(lr){5-6}\cmidrule(lr){7-8}\cmidrule(lr){9-10}\cmidrule(lr){11-12}\cmidrule(lr){13-14}
BS & Policy & T1g & T1p & T1g & T1p & T1g & T1p & T1g & T1p & T1g & T1p & T1g & T1p \\
\midrule
\multirow{4}{*}{4}
  & Proposed       & 1.063 & -- & 0.989 & -- & 0.972 & -- & 1.055 & -- & 0.993 & -- & 1.009$^\circ$ & -- \\
  & N-gram-first   & 1.069 & 1.042 & 0.980 & 0.989 & 0.981 & 1.008 & 1.047 & 1.013 & 0.960 & 0.977 & 1.013 & 1.016 \\
  & Extension      & 1.044 & 1.017 & 0.993$^\circ$ & 1.010 & 0.987$^\circ$ & 1.009 & 1.038 & 0.999$^\circ$ & 1.005$^\circ$ & 1.017 & 1.007$^\circ$ & 1.012 \\
  & \dspark (prob.) &  & 0.978 &  & 1.013 &  & 1.000$^\circ$ &  & 0.974 &  & 1.013 &  & 1.012$^\circ$ \\
\midrule
\multirow{4}{*}{128}
  & Proposed       & 0.997$^\circ$ & -- & 0.975 & -- & 0.981 & -- & 1.002$^\circ$ & -- & 0.979 & -- & 0.989 & -- \\
  & N-gram-first   & 0.978 & 0.949 & 0.965 & 0.951 & 0.974 & 0.956 & 0.995$^\circ$ & 0.978 & 0.961 & 0.954 & 0.982 & 0.971 \\
  & Extension      & 0.998$^\circ$ & 0.970 & 0.988 & 0.970 & 0.989$^\circ$ & 0.968 & 1.003$^\circ$ & 0.982 & 0.985 & 0.970 & 0.998$^\circ$ & 0.975 \\
  & \dspark (prob.) &  & 0.955 &  & 0.966 &  & 0.980 &  & 0.980 &  & 0.977 &  & 0.986 \\
\bottomrule
\end{tabular}}
\end{table}

\section{Stale budget sizing}
\label{app:stale}

This appendix expands \cref{sec:eval-stale}.

\textbf{Where staleness comes from.} \av splits its work across the boundary
of asynchronous scheduling (\cref{sec:background}). The scheduler on the host
fixes the size of the next verification pass, the budget $b^*$, before the GPU
has produced the survivals for that step, so it sizes the budget from survivals
one step old. The GPU then decides which (request, position) pairs get the
$b^*$ slots from the fresh survivals. This split is \dspark's
design~\citep[\S5.2]{cheng2026dspark}, and \sys keeps it: only the budget's size
and the slot price $\lambda$ are stale, never the allocation. A stale budget
is too small or too large by however much the batch's survivals changed in one
step.

\textbf{Why the n-gram drafter suffers more than \dspark.} \dspark proposes a
chain on every step, and its confidences drift only in magnitude from one step
to the next. An n-gram match instead appears and vanishes abruptly as the
output enters and leaves a region it copies from its input. A request that has
just entered a copy region has stale survivals with no match, so the host sizes
the budget as if the request were worth nothing and misses its 15-slot chain.
A request that has just left one is counted for a chain it no longer has. We
measure the per-request \emph{miscount} $d$, the expected accepted drafts
under the stale survivals minus under the fresh ones. For the n-gram drafter, a
match is gained or lost between the previous step and the current one for
5.7--9.2\% of requests in each direction, and these requests are miscounted by
$-4.4$ to $-5.3$ and $+5.4$ to $+7.1$ drafts. In \sys a request's chosen chain
switches between n-gram and \dspark instead, for 8.8--13.7\% of requests in
each direction, with miscounts of $-3.0$ to $-5.0$ and $+4.6$ to $+7.5$. The
standard deviation $\sigma$ of $d$ is 0.85--1.74 for \dspark, 2.2--3.0 for the
n-gram drafter and 2.4--3.4 for \sys. \sys's is the largest because it has
both sources of error: chain switches, as in the n-gram drafter, and drift in
the confidences of the \dspark chains that the other requests carry.

\textbf{The error grows as $\sqrt{B}$.} The batch miscount is the sum of the
per-request miscounts, $D = \sum_{i=1}^{B} d_i$. Two measured properties make
its size predictable. First, $\sigma$ does not depend on the batch size. For
each model, arm and workload, it varies by at most 0.10 across BS~4, 32 and
128, since how often a request's match comes and goes is a property of the
request, not of the batch. Second, requests miscount independently: the
standard deviation of $D$ is 0.96--1.11 times $\sigma\sqrt{B}$ in every run.
$D$ therefore has standard deviation $\sigma\sqrt{B}$ and, as a sum of many
independent terms, is close to normal with mean near zero. The budget is too
small when $D < 0$, and on those steps the shortfall is the mean of a
half-normal distribution:
\begin{equation}
\mathbb{E}\bigl[-D \mid D < 0\bigr] = \sqrt{2/\pi}\;\sigma\sqrt{B}
\approx 0.80\,\sigma\sqrt{B}.
\label{eq:halfnormal}
\end{equation}
The \textbf{expected accepted drafts of the batch grow as $\boldsymbol{B}$, so
the shortfall relative to them falls as $\boldsymbol{1/\sqrt{B}}$} (column
Rel.\ in \cref{tab:stale}).

\textbf{Methodology.} We instrument the engine to record, at every full-batch
step, both the stale survivals that sized the budget and the fresh ones. We
re-pick the budget from each with \av's own rule, which reproduces the budget
the engine chose on 100\% of decisions, and score both budgets on the fresh
survivals. Goodput lost is $1 - G_{\text{stale}} / G_{\text{fresh}}$, averaged
over steps. It uses the estimator's predicted acceptance, so only the
difference between stale and fresh is meaningful. We run \dspark, n-gram +
\av and \sys on Qwen3-4B and Llama-3.1-8B, on InstructCoder and NextCoder
15--20\% and 40--45\%, at BS~4, 32 and 128, each on one GPU (54 runs).

\begin{table}[H]
\centering
\small
\caption{The half-normal model predicts the stale budget's shortfall, and the
relative shortfall and its goodput cost fall with batch size. Qwen3-4B,
NextCoder 15--20\%. $\sigma$: standard deviation of the per-request miscount.
Pred.: \cref{eq:halfnormal}, from the measured $\sigma$ and the number of
requests with previous-step survivals (3.9--4.0, 31.3--31.7 and 125.5--127.0 on
average). Obs.: mean shortfall on steps where the stale budget is too small.
Both in expected accepted drafts. Rel.: Obs.\ divided by the batch's expected
accepted drafts. Lost: goodput lost (\%).}
\label{tab:stale}
\setlength{\tabcolsep}{3.5pt}
\begin{tabular}{lrrrrrr}
\toprule
Arm & BS & $\sigma$ & Pred. & Obs. & Rel. & Lost \\
\midrule
\multirow{3}{*}{\dspark}
  & 4   & 1.37 & 2.16  & 1.94  & 8.1\%  & 0.29 \\
  & 32  & 1.36 & 6.11  & 5.63  & 3.0\%  & 0.16 \\
  & 128 & 1.37 & 12.24 & 11.03 & 1.4\%  & 0.06 \\
\midrule
\multirow{3}{*}{n-gram + \av}
  & 4   & 2.98 & 4.74  & 4.23  & 39.4\% & 7.17 \\
  & 32  & 3.00 & 13.50 & 13.11 & 16.0\% & 1.82 \\
  & 128 & 2.96 & 26.60 & 28.04 & 8.5\%  & 0.35 \\
\midrule
\multirow{3}{*}{\sys}
  & 4   & 3.39 & 5.35  & 4.72  & 13.5\% & 2.63 \\
  & 32  & 3.42 & 15.26 & 14.78 & 5.3\%  & 0.80 \\
  & 128 & 3.43 & 30.66 & 30.73 & 2.8\%  & 0.30 \\
\bottomrule
\end{tabular}
\end{table}

\textbf{The prediction matches the measurement.} \Cref{tab:stale} compares
\cref{eq:halfnormal} with the measured shortfall. At BS~32 and 128 the
observed shortfall is 0.88--1.14 times the prediction on every NextCoder run of
both models, and 0.77--1.14 times including InstructCoder. At BS~4 a sum of
four terms is far from normal, and the ratio spreads from 0.80 to 1.36. For
every arm, the relative shortfall falls roughly as $1/\sqrt{B}$, from 8.1\% to
1.4\% for \dspark, 39.4\% to 8.5\% for the n-gram drafter and 13.5\% to 2.8\%
for \sys.

\textbf{The error is largest where \av matters least.} On Qwen3-4B across the
three workloads, stale sizing costs the n-gram drafter 7.2--8.9\% of goodput
at BS~4 and \sys 2.3--2.6\%, but at BS~128 it costs them at most 0.64\% and
0.30\% on NextCoder. At BS~4 verification is nearly free and \av has little to
gain. At BS~128, where \av gives the n-gram drafter its 1.90\x\ over verifying
every draft (\cref{sec:motivation}), staleness costs almost nothing. Every
throughput number in this paper runs \av at every batch size, so it already
includes this loss.

\textbf{A shortfall costs much less goodput than its size.} At BS~4, \sys's
budget falls short by 13.5\% of the batch's expected accepted drafts but loses
2.6\% of goodput, and the n-gram drafter's falls short by 39.4\% and loses
7.2\%. Three effects shrink the cost. First, the allocation is fresh, so a
short budget drops the lowest-ranked slots, whose survivals are closest to
$\lambda$ and which by \cref{eq:price} add the least goodput. A long budget
likewise adds slots just below $\lambda$. Second, verifying fewer slots also
costs less, so the goodput ratio partly corrects itself. Third, the budget is
too small on only about half the steps. \sys also loses less than the n-gram drafter alone. When a request finds a
match that its stale survivals miss, the host still budgets for the request's
\dspark chain rather than for nothing. The budget therefore falls short by
less, and the request receives 72--80\% of its due slots at BS~4, against
29--36\% for the n-gram drafter alone.

\section{Context length on SPEED-Bench}
\label{app:sbctx}

\begin{figure}[H]
  \centering
  \includegraphics[width=\textwidth]{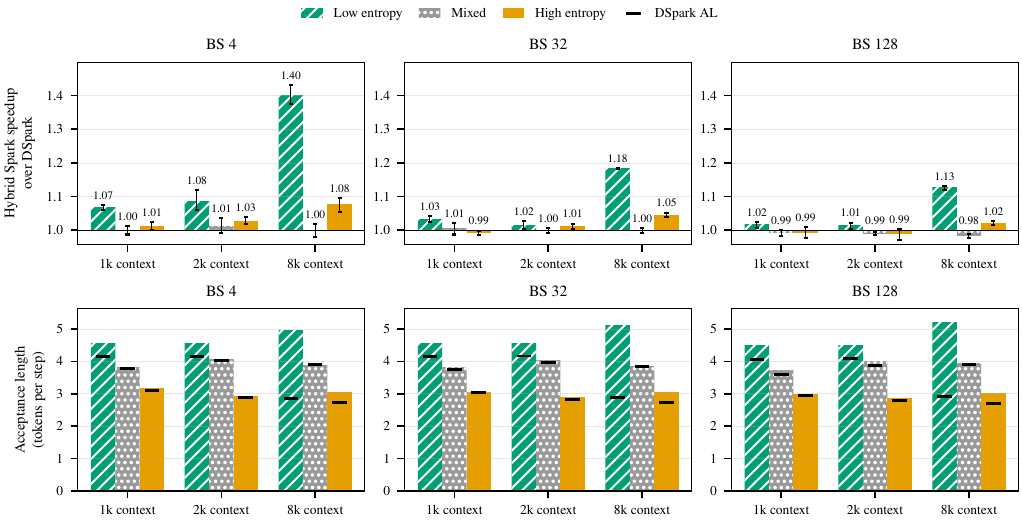}
  \caption{\sys's gain over \dspark grows with context length on low-entropy
  SPEED-Bench prompts, where \dspark's AL falls at 8k and \sys's rises, and
  stays level on mixed ones. Top: output-throughput speedup of \sys over
  \dspark (higher is better), with 95\% CIs. Bottom: AL of \sys (bars) and
  \dspark (black ticks). Qwen3-4B, T{=}0, FP8 KV cache, prompts not used to
  fit the estimator, mean of 3 repeats.}
  \label{fig:sbctx}
\end{figure}

On low-entropy prompts, \sys's gain over \dspark grows with context length, while on
mixed prompts the two are level. \Cref{fig:sbctx} shows the speedups and ALs
for Qwen3-4B at T{=}0 on SPEED-Bench prompts of 1k, 2k and 8k tokens.

\textbf{Prompts.} SPEED-Bench's throughput split has three categories, which we
run as separate workloads: low entropy (code completion and sorting), mixed
(question answering and needle-in-a-haystack) and high entropy (book
continuation and creative writing). The 8k split has 1{,}536 prompts, 512 per
category (1{,}530 distinct). The estimator is fit on 764 of them
(\cref{sec:fitting}), so we use only the other 766: 256 low-entropy, 256
mixed and 254 high-entropy. At 1k and 2k we drop prompts drawn from the same
source document as a fitting prompt and sample the same number per category as
at 8k. Each request generates up to 512 tokens. BS~4 and 32 send each prompt
once, BS~128 twice.

\textbf{Serving.} Every arm runs on two GPUs with tensor parallelism at every
batch size, with an FP8 KV cache, so that 128 requests with 8k-token prompts
fit in memory, and prefix caching off. FP8 lowers \dspark's AL slightly (2.7
against 2.8 with the BF16 KV cache on the 8k high-entropy prompts at BS~128),
so these speedups are not directly comparable with the rest of the paper. Each
cell is the mean of 3 repeats, each on a fresh server.

\textbf{Low entropy.} \sys's speedup over \dspark rises from 1.07\x at 1k to
1.40\x at 8k at BS~4, and from 1.02\x to 1.13\x at BS~128. As with EditBench
requests over 6{,}000 tokens of context (\cref{sec:eval-ood}), \dspark's AL
falls with context, from 4.1 at 1k to 2.9 at 8k, while \sys's rises from 4.6 to
5.0 at BS~4. The n-gram drafter copies from the longer prompt what \dspark misses. At 8k,
\sys is 2.35\x faster than vanilla decoding at BS~4 and 1.36\x at BS~128.

\textbf{Mixed and high entropy.} On mixed prompts the AL of both systems is
3.6--4.0 at every length, and \sys is level with \dspark at BS~4 and 32 and
0.98--0.99\x at BS~128. On high-entropy prompts the gain is small but also
grows with context, from 1.01\x to 1.08\x at BS~4 and from 0.99\x (level) to
1.02\x at BS~128.

\end{document}